\documentclass[journal]{IEEEtran}
\usepackage{cite} 

\usepackage[citecolor=blue,
linkcolor=blue,
unicode=true,
urlcolor=blue]{hyperref}
\usepackage{clrscode}
\usepackage{graphicx}
\usepackage{enumerate}
\usepackage{epsfig}
\usepackage[T1]{fontenc}    % use 8-bit T1 fonts
\usepackage{url}            % simple URL typesetting
\usepackage{amsfonts}       % blackboard math symbols
\usepackage{nicefrac}       % compact symbols for 1/2, etc.
\usepackage{microtype}      % microtypography
\usepackage{multirow}
\usepackage{subfig}
\usepackage{boxedminipage}
\usepackage{array}
\usepackage{rotating}
\usepackage{bbm}
\usepackage{float}
\usepackage{wrapfig,lipsum,booktabs}
\usepackage{amsmath, amssymb, amsthm, caption,bm}
\numberwithin{equation}{section}
\usepackage{algorithm}
\usepackage{algpseudocode}
\usepackage{ragged2e}

\newtheorem{remark}{Remark}[section]

\newtheorem{theorem}{$\bm{\mathrm{Theorem}}$}[section]

\def\sign{{\rm sign}}
\usepackage{bm}
\usepackage{verbatim}

\renewcommand{\theequation}{\arabic{equation}}

\ifCLASSINFOpdf
\else
\fi
\begin{document}
\title{Leveraging both Lesion Features and Procedural Bias in Neuroimaging: An Dual-Task Split dynamics of inverse scale space}

\author{
    \IEEEauthorblockN{Xinwei Sun \IEEEauthorrefmark{1}, Wenjing Han\IEEEauthorrefmark{2}, Lingjing Hu  \IEEEauthorrefmark{2}, Yuan Yao  \IEEEauthorrefmark{3}, Yizhou Wang \IEEEauthorrefmark{4}} \\
    \IEEEauthorblockA{\IEEEauthorrefmark{1}Microsoft Research Asia
    xinsun@microsoft.com} \\
    \IEEEauthorblockA{\IEEEauthorrefmark{2}Yanjing Medical College, Captial Medical University
    \{hanwj2014,hulj\}@ccmu.edu.cn }  \\
    \IEEEauthorblockA{\IEEEauthorrefmark{3} Hong Kong University
of Science and Technology. 
    yuany@ust.hk} \\
    \IEEEauthorblockA{\IEEEauthorrefmark{4}Center on Frontiers of Computing Studies,
Adv. Inst. of Info. Tech, 
Dept. of  Computer Science, Peking University
    Yizhou.Wang@pku.edu.cn} \\
    \thanks{Corresponding author: Lingjing Hu (email: hulj@ccmu.edu.cn), Yuan Yao (email: yuany@ust.hk)}
}

\markboth{Journal of \LaTeX\ Class Files,~Vol.~14, No.~8, August~2015}%
{Shell \MakeLowercase{\textit{et al.}}: Bare Demo of IEEEtran.cls for IEEE Journals}



\maketitle

\begin{abstract}
The prediction and selection of lesion features are two important tasks in voxel-based neuroimage analysis. Existing multivariate learning models take two tasks equivalently and optimize simultaneously. However, in addition to lesion features, we observe that there is another type of features, which are commonly introduced during the procedure of preprocessing steps, can improve the prediction result. We call such a type of features as \textbf{\emph{procedural bias}}. Therefore, in this paper, we propose that the features/voxels in neuroimage data are consist of three orthogonal parts: lesion features, procedural bias and null features. To stably select lesion features and leverage procedural bias into prediction, we propose an iterative algorithm (termed \textbf{GSplit LBI}) as discretization of differential inclusion of inverse scale sapce, which is the combination of \emph{Variable Splitting scheme} and \emph{Linearized Bregman Iteration (LBI)}. Specifically, with a variable splitting term, two estimators are introduced and split apart, i.e. one is for feature selection (the sparse estimator) and the other is for prediction (the dense estimator). Implemented with Linearized Bregman Iteration (LBI), the solution path of both estimators can be returned with different sparsity level on the sparse estimator for selection of lesion features. Besides, the dense estimator can additionally leverage procedural bias to further improve prediction results. To test the efficacy of our method, we conduct experiments on simulated study and Alzheimer's Disease Neuroimaging Initiative (ADNI) database. The validity and benefit of our model can be shown by the improvement of prediction results and interpretability of visualized procedural bias and lesion features. 
\end{abstract}


% \fi
\IEEEpeerreviewmaketitle
\setlength\parindent{0pt}

\section{Introduction}
\IEEEPARstart{A}{ccurate} prediction of cognitive or disease state is a major goal in neuroimage analysis. One can apply univariate models (e.g. two-sample T-test) to determine which voxels show significant correlations with the task, such as statistical parametric maps (SPMs) \cite{SPM}. However, the voxel-independence assumption behind such process fails to capture the spatial correlation across different brain regions. Therefore, multivariate approaches in recent years gained more attention in neuroimaging area \cite{Multi-apply-svm,Multi-apply-fmri} and proved to be superior than univariate models in terms of prediction power and interpretability \cite{Multi,Multi-uni}, e.g., SVM \cite{Multi-apply-svm}. 

In addition to disease prediction, the stable selection of lesion voxels (disease-relevant) is also of concern for neuro-scientists since they can be used to interpret the disease, as a clue for diagnose and further exploration of potential mechanism with therapy \cite{neuropathologic}. For example, Alzheimer's Disease (AD), it's commonly believed that the regions suffer from degenerations are located in parahippocampus gyrus, hippocampus gyrus, temporal lobe, parietal lobe, cingulate gyrus, etc. \cite{grey_matter, hippocampus_3d, neuropathologic, 4d}. 

%To achieve this goal, one can extract regions of interest (ROI), i.e., pre-defined regions with each corresponding to a neuroanatomical organization, as inputs for classification. However, the pre-definition of regions excludes the possibility when only a few voxels in the whole region are correlated with the disease label. To overcome such a drawback, another alternative approach is voxel-based analysis, in which the input features are the value of voxel volume (e.g. gray matter volume). Each voxel is embedded in a 3-dimensional coordinate position. 

In this paper, we consider using voxel-based analysis, which provides finer scale than Region of interest (ROI)-based method and hence has been increasingly applied  \cite{VBM-morph,VBM-ad} with the implementation of structural Magnetic Resonance Imaging (sMRI) that brings no harm to human brains. Due to collection cost, the neuroimage data always suffer from the limited (e.g. dozens of or hundreds of) observations with much higher dimensional feature space (e.g. thousands or millions of features). To avoid over-fitting, existing multivariate machine learning has been applied with regularization function to enforce the pattern among features. Specifically, for neuroimage disease considered in this paper, it is commonly hypothesized that lesion voxels are the most interpretable features to the disease with the following structural sparsity: (i) sparsity: only a subset of features can be regarded as lesion features (ii) geometrically smoothness: each voxel shares the similar extent of degeneration/lesion with its neighbors; (iii) non-negative correlation with the disease: e.g., in Alzheimer's Disease, the gray matter voxel among patients tends to be atrophied rather than enlarged, compared with that among normal controls.

In the literature, existing models take the task of selection of lesion features and prediction of disease equivalently in terms of optimization. Specifically, one corresponding strategy that often adopted by these machine learning models is that: among the pre-setting grid of regularization parameters that reflects the sparsity levels of selected features, the ones corresponding to the highest prediction result are applied to the model for feature selection. The form of regularization scheme largely depends on the prior knowledge of the underlying structure of the disease-relevant features which may lead to different regularization forms. For instance, lasso \cite{lasso} assumes sparsity; elastic net \cite{elastic} considers additional property of grouping correlated features; group lasso manually define grouping of features \cite{group1,group2,group3,group4} and particularly TV-$L_{1}$ (Combination of Total Variation \cite{tv,nonlinear-tv} and $L_{1}$), takes into account both spatial cohesion and sparsity \cite{tv,tv-l1,efficient}. 

\begin{figure*}
\centering
\begin{minipage}[t]{0.42\linewidth}
    \includegraphics[width= \columnwidth]{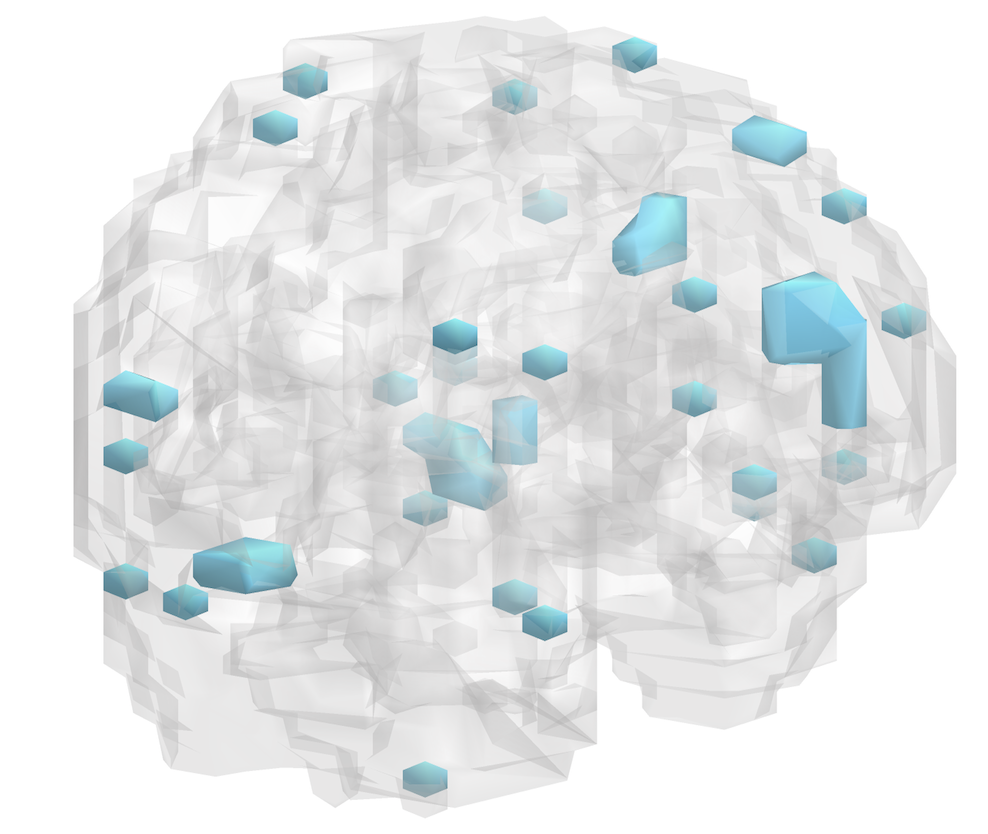}%{\begin{center} ori \end{center}}
\end{minipage}    
\begin{minipage}[t]{0.54\linewidth}
    \includegraphics[width= \columnwidth]{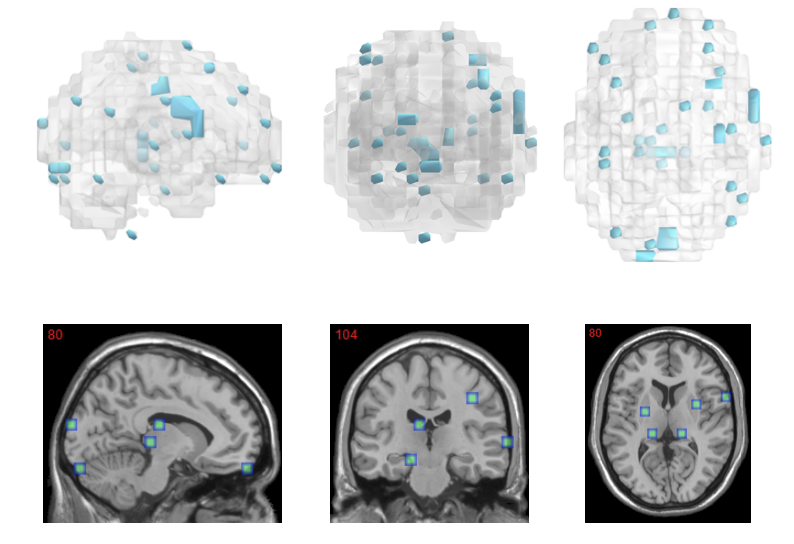}%{\begin{center} ori \end{center}}
\end{minipage}   
\caption{The top 50 negative voxels of average $\beta_{\mathrm{pre}}$ at the time corresponding to the highest accuracy in the path of GSplit LBI using 8-fold cross-validation. For subjects with AD, they represent enlarged GM voxels surrounding lateral ventricle, subarachnoid space, edge of gyrus, etc.}\label{figure:procedural-bias}
\end{figure*}

However, as pointed out by George Box in \cite{Box1979Robustness}, \textbf{"\emph{all models are wrong}"}, we observe that it's not reasonable to take these two tasks (selection of lesion features, prediction) equivalently in voxel-based neuroimage analysis. One major component of the gap in neuroimage analysis here is dominated by some fixed patterns of biases introduced during the preprocessing of $\mathrm{T_{1}}$-weighted image, i.e. the first step of voxel-based neuroimage analysis, such as segmentation and registration of grey matter (GM), white matter (WM) and cerebral spinal fluid (CSF). Such biases are due to scanner difference, different population and especially limitation of spatial normalization, etc. \cite{whyvbmused}. Part of them can be helpful to the discrimination of subjects from normal controls (NC), but may not be directly related to the disease. For example in sMRI of subjects with Alzheimer’s Disease (AD), after spatial normalization during simultaneous registration of GM, WM and CSF, the GM voxels surrounding lateral ventricle and subarachnoid space etc. may be mistakenly enlarged caused by the enlargement of CSF space in those locations \cite{whyvbmused} compared to the normal template, as shown in Fig.~\ref{figure:procedural-bias}. Although these voxels/features are highly correlated with disease, they cannot be regarded as lesion features in an interpretable model, in this paper we refer to them as "Procedural Bias", which should be identified
but are neglected in the literature. We observe that it can be explicitly harnessed in our voxel-based image analysis to improve the prediction of disease.

The existence of procedural bias makes the prediction a different goal from selection of lesion features. Existing models fail to take both tasks into consideration. Specifically, as mentioned earlier, models with regularization such as TV-L$_{1}$ and $n^{2}$GFL \cite{n2gfl}, can enforce strong prior of disease in order to capture the lesion features. Although such features are disease-relevant, these models lose some prediction power without consideration of the procedural bias. To pursue a better prediction power, one may utilize the information in procedural bias. For example, the model with $\ell_2$-based regularization (e.g. ridge regression \cite{Ridge-linear}, elastic net \cite{elastic} and graphnet \cite{graphnet}) select strongly correlated features to minimize classification error. However, note that the procedural bias is different from lesion voxels in terms of volumetric information (enlarged v.s. degenerated) and spatial pattern (surrounding distributed v.s. spatial cohesive). Therefore, such regularizations without differentiation of procedural bias from lesion features suffer from poor interpretability and hence may be prone to over-fit. 

%caused by the disease; besides, the mixture of two kinds of features can violate the differentiation of lesion ones from procedural bias and be prone to leading an uninterpretable model. Therefore, it may not be reasonable to mix two tasks (prediction and selection of lesion features) in neuroimage analysis. In addition, it's hard for a single model to take both into consideration. Roughly speaking, existing models can be categorized into two classes: there are mainly two classes of models in literature, both of which fail in utilizing the information in procedural bias effectively. 

Motivated by the fact that procedural bias is different from pathology-related lesion features, we in this paper propose that the voxels in the whole brain are orthogonally decomposition of three kinds of voxels: (i) atrophied voxels that are contributed to the disease, i.e. lesion features that are interpretable to the disease; (ii) the mistakenly enlarged voxels, i.e. procedural bias which is correlated with the disease however is not belong to the lesion ones; (iii) other voxels that have ignorable correlations with the disease, i.e. null features.

To fullfill the two different tasks that (i) leverage procedural bias into prediction and (ii) stably selection lesion features, we appended the loss function as negative log-likelihood of \emph{g}eneralized linaer model (GLM) with variable \emph{splitting} term and implement dynamics training in \emph{inverse scale space} induced by \emph{Linearized Bregman} (LB) Distance, namely \emph{G}eneralized \emph{Split}ting \emph{L}inearized \emph{B}regman \emph{I}nverse \emph{S}cale \emph{s}pace (GSplit LBISS). Specifically, we adopted the variable splitting scheme \cite{splitlbi,Ye2011Split,Bo2012An}: the original parameter space $\beta_{\mathrm{pre}}$ is lifited to a coupled pair $(\beta_{\mathrm{pre}},\gamma)$ constrained by $\ell_2$ regularization: with sparse estimator $\gamma$ selecting lesion features and dense estimator $\beta_{\mathrm{pre}}$ to additionally capturing procedural bias for better prediction. For training, we proposed a dynamic approach, which generates a family of paired estimators respectively for the above two tasks as solutions in \emph{differential inclusion of inverse scale space}: $\gamma$ learns the structural sparsity in inverse scale space and $\beta_{\mathrm{pre}}$ that is dense estimator to exploit the procedural bias and lesion features for better prediction. The GSplit LBISS enjoys a simple discretization, namely \emph{G}eneralized \emph{Split}ting \emph{L}inearized \emph{B}regman \emph{I}teration  (GSplit LBI). At each step, the sparse estimator is returned as the projection of the dense estimator onto the subspace of the support set of $\gamma$, satisfies the structural sparsity and hence can be used for selection of lesion features. The remainder of this projection is heavily influenced by the procedural bias, which can be captured by the dense estimator. 

Historically, the Linearized Bregman Iteration (LBI) was firstly proposed as sparse recovery optimization algorithm in image denoising \cite{osher2005an} and compressed sensing. In \cite{osher2016sparse, huang2018unified} firstly consider the LBI as discretization of differential inclusion and presented its statistical property in sparse signal recovery in high-dimensional space. Specifically, under the same condition with Lasso, the LBI can enjoys model selection consistency. In this paper, we generalize it to generalized linear model and structural sparsity, hence the consistent property in feature selection is also inherited. In addition, equipped with variable splitting term that endows the dense parameter some degree of freedom to capture the procedural bias. Therefore, our GSplit LBI can enjoy interpretable feature selection of lesion features and meanwhile the ability of leverage procedural bias to improve prediction power. Besides, our algorithm is easy to implement and is much more efficient than existing sparsity models (e.g. lasso, generalized lasso \cite{genlasso}).

%It should be noted that the GSplit LBI is the discretization of the differential inclusion. Hence, the model selection consistency property can be discussed, which means that equipped with an early stopping scheme, our algorithm can select true lesion features. Moreover, the dense estimator can additionally leverage procedural bias to improve prediction result and interpretability. Our algorithm is easy to implement and is much more efficient than existing sparsity models (e.g. lasso, generalized lasso \cite{genlasso}.  

To test the validity, in this paper we apply our model to voxel-based structural MRI (sMRI) analysis for Alzheimer's Disease (AD), which is challenging and attracts increasing attention lately. In detail, equipped with a few reasonable prior knowledge on lesion features of AD, we apply GSplit LBI to the classification of AD, Mild Cognitive Impairment (MCI) and NC. Fitting GSplit LBI on totally 546 samples, with the capture of additional procedural bias, we achieve comparable or better prediction results than state-of-art methods. On the other hand, the selected lesion features are the most stable among all listed models. Moreover, they are stably concentrated in locations such as Hippocampus, ParaHippocampal gyrus and medial temporal lobe, which are believed to be the early damaged regions. Finally, we discuss the differentiation of procedural bias and lesion features by GSplit LBI, future improvements and extend applications of our methods on finer scales. 

This paper is the extended version of the MICCAI'2017 conference report \cite{sun2017gsplit} which only gives a brief introduction of GSplit LBI algorithm with preliminary experimental results on cross-validation datasets. In this paper, we extend the report above in both methodology and experiments. In methodology, we give a detailed development of GSplit LBI, as discretization of differential inclusion of inverse scale space, its connection to mirror descent algorithm \cite{nemirovsky1983problem} and computational complexity. In experiments, we conduct additional simulation experiments to illustrate the ability of proposed method to capture both lesion features and procedural bias, and additional classification results of MCI and NC on dataset processed by 3.0T field strength Magnetic Resonance Imaging scanner. Moreover, the results are tested based on held-out datasets, which is a statistically better evaluation of prediction power than merely the cross-validation.

%I wish you the best of success.

 

\subsection{Paper Organization}
The paper is organized as follows. In Section~\ref{sec:method}, we introduce our loss function for classification and also the structural sparsity priori that the our solution should satisfies. Then we introduce a dual-task differential inclusion of inverse scale space, namely GSplit LBISS and its discretization form called GSplit LBI, including properties variable splitting, regularization solution path and computational complexity; The simulation experiment is introduced and discussed in Section~\ref{sec:simulation}, implying that our algorithm can capture both procedural bias and lesion features; Section~\ref{sec:adni} conducts experimental results on both disease prediction and lesion features analysis. Besides, we further conduct coarse-to-fine experiment to investigate the locus of selected lesion features. The detailed discussion of experimental results and conclusions are presented in Section~\ref{sec:discussion}.

\section{Related Work}
\label{sec:relate}

\subsection{Early Prediction of Neurological Disease}
\label{sec:early-prediction}

Accurate pre-diagnosis of brain diseases, especially those irreversible ones, is significant for early medical intervention and treatment. Many neurodegenerative diseases can suffer from structural damages for decades before onset of clinical symptoms, which provides a window for early prediction in terms of abnormal brain regions. For example, there could be progressively atrophy in regions such as hippocampus, medial temporal lobe and amygdala, etc. in Alzheimer's Disease (AD) \cite{braak1991neuropathological, visser2002medial, dubois2007research}. To capture these biomarkers for accurate prediction, one can feed into machine learning models preprocessed data on images acquired from Structural Magnetic resonance Imaging (sMRI) (e.g.. T$_1$-weighted image), which have been shown to be effective in finding morphological changes \cite{kloppel2012diagnostic}, as a non-invasive imaging to human brains. For example, one can preprocess sMRI data to extract gray matter (GM) voxels, which suffered loss in volume in diseases such as AD, Frontotemporal dementia \cite{whitwell2010gray}, Corticobasal degeneration \cite{boxer2006patterns}, Parkinson's Disease \cite{xia2013changes}. Together with statistical inference on preprocessed GM voxels for abnormality analysis, i.e.,  voxel-based morphometry (VBM), it has been shown effective for prediction for AD \cite{VBM}. 

However, some biases are inevitably introduced during the procedure of pre-processing steps such as registration, segmentation and modulation, especially in aging patients' brains \cite{kennedy2009age}, leading to false-discovery results in VBM. Such biases are mainly due to abnormalities in patients brain, and hence can be regarded as additional signals for classification between normal controls and patients. Again taking AD as an example, as aforementioned, these biases correspond to mistakenly enlarged GM voxels caused by enlarged CSF space in lateral ventricles \cite{duran2006risk} and subarachnoid spaces, etc. This type of features, which is namely \emph{procedural bias} in this paper, is the first time to be proposed to help prediction. We applied it to AD, which is irreversible and on the rise recently,  and have shown improved prediction accuracy. 

\subsection{Feature Selection}
\label{sec:feature}
In addition to prediction, the interpretability of selected biomarkers/features is another important issue in clinical practice. The feature selection methods can be roughly categorized as two classes: univariate analysis and multivariate analysis. The most typical method in univariate class is two-sample T-test, in which we independently implement hypothesis test voxel-wise, which suffers from high false-discovery-rate (FDR). Other methods, such as BH$_q$ \cite{benjamini1995controlling}, localfdr \cite{efron2005local}, FDR-HS \cite{sun2018fdr} can alleviate FDR problem, however they treat features independently and hence cannot capture the underlying structure such as spatial cohesion of features. 

Multivariate analysis, as an alternative to univariate analysis, is more powerful in utilizing correlations among features and has been successfully utilized in neuroimage analysis. As a sparse feature selection method, the Lasso selects a subset of voxels are correlated with the disease. It ever achieved one of the state-of-the-arts in brain image classification \cite{lasso-apply}, however suffers from low stability of lesion features selection \cite{n2gfl}. To overcomes this problem, the elastic-net \cite{elastic} combines the effect of sparsity enforced by lasso-type penalty and strict convexity enforced by ridge-type penalty. In contrast to lasso, it can select clustered area rather than single scattered voxels \cite{wang2017elastic}. To further consider the spatial cohesion of voxels which are embedded in a 3-dimensional space (e.g., atrophied voxels form in cluster such as hippocampus), one can implement Total-Variation (TV)-type sparsity \cite{tv} which enforces correlation between voxels and their neighbors. Compared with group lasso penalty which allows predefined groups of voxels to be selected or not simultaneously, the TV can explores new damaged regions rather than pre-defined ones. Particularly, \cite{efficient} combined lasso-type penalty and TV-type penalty and thus can select atrophied regions. Further, the \cite{n2gfl} additionally enforced the positive correlation between the lesion voxels and the disease label, which is reasonable in many neuroimage analysis (such as AD, demential with Lewy bodies, Parkinson Disease, etc), and achieved more stable result in terms of feature selection and prediction. However, these multivariate analysis suffered from two problems: (i) the multi-collinearity problem \cite{tu2005problems} which means high correlation among features and hence can select spurious correlations, especially in high-dimensional analysis (ii) failed to consider the aforementioned procedural bias that can improve classification result. In this paper, we proposed to resolve these two problems to achieve better interpretability and prediction power.

%\begin{figure}[!h]
%\centering
%\includegraphics[width= 0.8\columnwidth]{image_pami/elasticnet.eps}
%\caption{Solution path from left to right: Lasso, Elastic Net ($\alpha = 0.4$) and Ridge. Each color represents solution path of each variable.}
%\label{elastic}
%\end{figure}

%Unlike $l_{1}$-norm which only selects one or few voxels in any regions, the $D_{G}\beta$ enforces correlation between the voxel and its neighbors. Together with lasso-type penalty, i.e. $\Omega(\beta) =  \lambda( \Vert \beta \Vert_{1} + \rho \Vert D_{G} \beta \Vert_{1})$, the sparsely clustered estimator can be returned, where $\rho > 0$ is the trade-off between geometric clustering and voxel sparsity. 

\section{Methodology}
\label{sec:method}

 \textbf{Problem Setup}  Our goal is to learn predictor $f: \mathcal{X} \to \mathcal{Y}$ that classifies the disease label $y \in \mathcal{Y}$ for any $x \in \mathcal{X}$ that collects the neuroimaging data with $p$ voxels. For disease/normal classification, $\mathcal{Y} := \{\pm 1\}$ ($+1$ denotes the normal control and $-1$ denotes the disease). To achieve this goal, we are given $N$ training samples $\{\mathbf{x}_{i},\mathbf{y}_{i}\}_{1}^{N}$. We denote $\mathbf{X} \in \mathbb{R}^{N \times p}$ and $\mathbf{y} \in \mathbb{R}^{p}$ as  concatenations of $\{\mathbf{x}_{i}\}_{1}^N$ and $\{\mathbf{y}_{i}\}_{1}^N$. In the rest, we will consider Alzheimer's Disease as an example to explain our method, whereas with the belief that the whole methodology can also be applied to other neuroimage diseases. 

\subsection{Lesion Features and Procedural Bias for Prediction}

%In this section, we systematically introduce our methodology in the scenario of voxel-based neuroimage analysis. Specifically, we start from introducing the generalized linear model for classification, followed by discussion of lesion features and procedural bias. We suggest that neuroimaging features/voxels consist of lesion features, procedural bias and null features. To capture non-null features (lesions and procedural bias), we introduce GSplit LBI algorithm, which incorporates most common types of structural sparsity in neuroimage analysis. Finally, we discuss the computational complexity at the end of this section. 

%\subsection{Generalized Linear model}
%\label{sec:glm}
%Our dataset consists of $N$ samples $\{\mathbf{x}_{i},\mathbf{y}_{i}\}_{1}^{N}$ where $\mathbf{x}_{i} \in \mathbb{R}^{p}$ collects the $i^{th}$ neuroimaging data with $p$ voxels and $\mathbf{y}_{i} = \{\pm 1\}$ indicates the disease status ($-1$ for disease in this paper). We denote $\mathbf{X} \in \mathbb{R}^{N \times p}$ and $\mathbf{y} \in \mathbb{R}^{p}$ as  concatenations of $\{\mathbf{x}_{i}\}_{1}^N$ and $\{\mathbf{y}_{i}\}_{1}^N$. Commonly speaking, the response variable $\mathbf{y}$ is generated from 
\textbf{Generalized Linear Model for Prediction} Consider the following discriminative loss for supervised learning: 
\begin{align}
\label{eq:discri}
\mathcal{L}_{\theta} = \mathbb{E}_{(x,y) \sim p(x,y)} -\log{p_{\theta}(y|x)}, 
\end{align}
in which $y|x$ is assumed to be generated from \emph{generalized linear model} (GLM): 
\begin{subequations} 
\label{eq:data-model}
\begin{align} 
\label{eq:glm-1}
 p(\mathbf{y}_i | \mathbf{x}_i,\beta_{\mathrm{pre}}^{\star},\beta_{\mathrm{pre},0}^{\star}) &  \propto   \mathrm{exp}\left(\frac{
\mu_{i}^{\star} \cdot y_i - \psi(\mu_{i}^{\star})}{d(\sigma)} \right)   \\
 \label{eq:mu}
  \mu_{i}^{\star}  = \langle \mathbf{x}_i, \beta_{\mathrm{pre}}^{\star} \rangle & + \beta_{0}^{\star}, \ \forall \thinspace i \in \{1,...,N\}
\end{align}
\end{subequations}
where $\beta_{\mathrm{pre}}^{\star}$ ($\beta_{0}^{\star}$) is true (intercept) parameter, $\psi: R \to R$ is link function and $d(\sigma)$ is known parameter related to the variance of distribution, and $\mu^{\star}_{i} := \langle \mathbf{x}_i, \beta^{\star} \rangle + \beta_{0}^{\star}$. With Eq.~\eqref{eq:data-model}, the $\theta := \{\beta_{0},\beta_{\mathrm{pre}}\}$ and the negative log-likelihood for sample $(x,y)$ is $\ell(\beta_{\mathrm{pre}},\beta_0|x,y) := -\log{p_{\theta}(y|x)}$. Given training data $(x_1,y_1),...,(x_N,y_N) \overset{i.i.d}{\sim} p(x,y)$, our goal turns to minimize the \emph{empirical risk minimization} (ERM), i.e., 
\begin{align}
\label{eq:erm}
\hat{\mathcal{L}}(\beta_{\mathrm{pre},\beta_0}) = \frac{1}{N} \sum_{i=1}^{N} \ell(\beta_{\mathrm{pre}},\beta_0|\mathbf{x}_i,\mathbf{y}_i).
\end{align}
For our binary classification task, the Eq.~\eqref{eq:discri} degenerates to commonly used logistic regression loss:
\begin{align}
\label{eq:logistic}
\hat{\mathcal{L}}(\beta_{\mathrm{pre}},\beta_0) = \frac{1}{N} \sum_{i=1}^{N} \left( \log{(1 + \exp(\mu \cdot \mathbf{y})) - \mu \cdot \mathbf{y}} \right). 
\end{align}

\textbf{Lesion Features}  In voxel-based neuroimage analysis, the lesion features have been the main focus since it provides the pathological criterion for the disease. The stable selection of lesion voxels can be helpful for better understanding of the disease, diagnose clinically and drug development. A desired estimator $\beta_{\mathrm{pre}}$ should not only minimize the loss function (Eq~\eqref{eq:erm}), but also satisfy the structural sparsity requirements, which are often enforced by regularization penalty $\Omega(\beta_{\mathrm{pre}})$. In this regard, the loss function turns to
\begin{align}
\beta_{\mathrm{pre}} = \arg \min_{\beta_{\mathrm{pre}}} \mathcal{L}(\beta_{\mathrm{pre}},\beta_0) + \Omega(\beta_{\mathrm{pre}}). \nonumber 
\end{align}
Such lesion features should satisfy the following \emph{priori} structural sparsities, with details explained in the later:
\begin{enumerate}
\item $\Vert \beta_{\mathrm{pre}} \Vert_1$: $\ell_1$ sparsity which implies that only a subset of features are related to the disease. 
\item $\Vert D_G \beta_{\mathrm{pre}} \Vert_1$: Total-Variation (TV) Sparsity, implying that voxel activities should be geometrically clustered or 3D-smooth. The $D_G : \mathbb{R}^V \to \mathbb{R}^E$ denotes a graph difference operator on $G = (V,E)$ \footnote{$V$ is the node set of voxels, $E$ is the edge set of voxel pairs in neighbor on 3-d space, i.e. $D_G\beta_{\mathrm{pre}} = \sum_{(i,j) \in E} (\beta_{\mathrm{pre}}(i) - \beta_{\mathrm{pre}}(j))$}.
\item $\mathbbm{1}(\beta_{\mathrm{pre}} \geq 0)$: Non-negative Correlation between disease status and volume of voxels.
\end{enumerate}
For 1), it was well known that only a subset of regions are strongly (causally) related to disease status for AD, such as medial temporal lobe and two-side hippocampus \cite{braak1991neuropathological, visser2002medial, dubois2007research}. The \cite{liu2012ensemble} has applied Lasso to AD and outperformed SVM. The 2) is due to the spatial coherence of lesion voxels, which means that these lesion features are geometrically clustered into regions such as two-side hippocampus and medial temporal lobe. Enforced by TV-type penalty, the prediction performance can be improved, with corresponding selected lesion features are located into early damaged regions for AD \cite{efficient}. Further, guided by another prior knowledge that the lesion features are often degenerated for many neuroimage diseases (such as AD, fron-totemporal dementia, corticobasal degeneration, etc.) \cite{n2gfl}, the 3) can further help improve stability of selection of lesion features and prediction accuracy. 

\textbf{Procedural Bias as a ``gap" between prediction and lesion voxels} In addition to lesion features, there is another type of features referring some fixed patterns of biases that are correlated with the disease, such as mistakenly enlarged gray matter voxels due to enlargement of neighbouring CSF space, as aforementioned in introduction. These biases are introduced during the prcedure of pre-processing step, such as spatial normalization and modulation step in \emph{V}oxel-\emph{B}ased \emph{M}orphometry (VBM) analysis for Alzheimer's Disease, hence we name it as \emph{procedural bias}. The procedural bias, which has not been taken into consideration for prediction in the literature, commonly exists as a gap between the selection of lesion features and prediction. In the subsequent section, we are going to introduce our method, namely \emph{G}eneralized \emph{Split} \emph{L}inearlized \emph{B}regman \emph{I}teration (GSplit LBI), as a dual-task method to handle prediction and stable selection of lesion features simultaneously.

\subsection{Dual-task dynamics via Differential Inclusion of Inverse Scale Space}
\label{sec:gsplit_lbi}
We introduce two parameters into our model, with the dense parameter for prediction that leverages procedural bias; and the sparse one for selecting lesion features. Such two parameters are enforced via a variable splitting scheme. Specifically, we adopt the variable splitting idea in \cite{splitlbi}, by introducing an auxiliary variable $\gamma$ enforced by sparse regularization $J(\gamma)$ to achieve priori structural sparsity 
requirements of lesion features. Meanwhile we keep it close to $D\beta_{\mathrm{pre}}$ via $\ell_2$ regularization: $S_{\rho}(D\beta_{\mathrm{pre}},\gamma) := \Vert D\beta_{\mathrm{pre}} - \gamma \Vert_{2}^2$, so as to inherit correlation existed in data between covariates $x$ and label $y$, supervised by training loss for $\beta_{\mathrm{pre}}$. Combined with Eq.~\eqref{eq:erm}, the updated loss is 
\begin{align}
    \label{eq:loss}
    \hat{\mathcal{L}}(\beta_{\mathrm{pre}},\beta_0, \gamma) = \hat{\mathcal{L}}(\beta_{\mathrm{pre}},\beta_0) + \frac{1}{2\nu} \Vert D\beta_{\mathrm{pre}} - \gamma \Vert_{2}^2,
\end{align}
where $\nu > 0$ controls the difference between $\gamma$ and $D\beta_{\mathrm{pre}}$, The $D = [I, \rho D_{G}^{\top}]^{\top}$ is the concatenation of identity matrix and the Laplacian matrix on graph of voxels embedded in 3d coordinate space. Correspondingly, we can correspondingly view $\gamma = [\gamma_V^\top , \gamma_G^\top]^\top$ as concatenation of $\gamma_V$ and $\gamma_G$. We can then split $\Vert D\beta_{\mathrm{pre}} - \gamma \Vert_2^2 $ into two terms: $\Vert D\beta_{\mathrm{pre}} - \gamma \Vert_2^2 = \Vert \beta_{\mathrm{pre}} - \gamma_V \Vert_2^2 + \Vert \rho D_G \beta_{\mathrm{pre}} - \gamma_G \Vert_2^2$, which can be explained as the combination of $\ell_1$ sparsity and TV sparsity, with $\rho > 0$ controlling the sparsity of lesion voxel and geometrically clustered. In addition, we can enforce $\gamma_V \geq 0$ to incorporate prior of non-negative correlation between disease label $y$ and lesion voxels. Our goal is thus two folds: (i) we use $\gamma$ as guidance for selection of lesion features and the structural sparsity of $\gamma$ can guide $\beta_{\mathrm{pre}}$ to learn corresponding sparsity pattern of lesions on corresponding support set of $\gamma$, and (ii) use dense parameters $\beta_{\mathrm{pre}}$ for prediction since it can additionally capture procedural bias located on other positions. 

To achieve the structural sparsity of $\gamma$ and additionally capture procedural bias inherited in the data, we consider the following differential inclusion: 
\begin{subequations}
\label{eq:gsplit-ode}
\begin{align}
\dot{\beta}^t_{0} & = -\nabla_{\beta^t_{0}} \hat{\mathcal{L}}(\beta^t_{\mathrm{pre}},\beta^t_0,\gamma^t) \label{eq:gsplit-ode-a} \\
\dot{\beta}^t_{\mathrm{pre}} & = -\nabla_{\beta^t_{\mathrm{pre}}} \hat{\mathcal{L}}(\beta^t_{\mathrm{pre}},\beta^t_0,\gamma^t) \label{eq:gsplit-ode-b} \\
\dot{v}^t & = -\nabla_{\gamma^t} \hat{\mathcal{L}}(\beta^t_{\mathrm{pre}},\beta^t_0,\gamma^t) \label{eq:gsplit-ode-c} \\
v^t & \in \partial \mathbf{J}(\gamma^t)  \label{eq:gsplit-ode-d},
\end{align}
\end{subequations}
where $v^0 = 0^{p+m}, \beta^0_{0} = 0, \beta^0_{\mathrm{pre}}=0^{p}, \gamma^0 = 0^{p+m}$, $\mathbf{J}(\gamma) = \Vert \gamma \Vert_1 + \frac{1}{2\kappa} \Vert \gamma \Vert_2^2 + \mathbbm{1}(\gamma_V \geq 0)$ and $v = \rho + \frac{\gamma}{\kappa} + \lambda$ ($\kappa > 0$) with $[-1,1] \ni \rho \in \partial \Vert \gamma \Vert_1, 0 \geq \lambda \in \partial \mathbbm{1}(\gamma_V \geq 0)$. The Eq.~\eqref{eq:gsplit-ode} generates a regularization solution path from simple to complex, starting from initial points $v^0 = 0$. Specifically, the $v^t = [v^{t,\top}_V,v^{t,\top}_G]^{\top}$ iterates as gradient descent flow in dual space, until hitting the $\ell_{\infty}$-unit box for $v^t_G$ or hitting 1 for $v^t_V$, implied by the fact that $v^t_i \in [-1,1] \leftrightarrow \gamma_i^t = 0$ for $i \in \{|V|+1,...,|V|+|E|\}$; and $v^t_i \geq 1 \leftrightarrow \gamma_i^t = 0$ for $i \in \{1,...,|V|\}$. Such a hit makes the corresponding $\gamma^t_G$ and $\gamma^t_V$ popping up to non-zero. The earlier the $v^t_G$ ($v^t_V$) reaches the $\ell_{\infty}$-unit box (or 1), the earlier of corresponding elements in $v^t_G$ (or $v^t_V$) selected to be non-zeros, which is called the \emph{inverse scale space} property, with "scale" meaning the selection order of elements in $\gamma$. When $\mathcal{L}$ denotes the squared loss (corresponds to linear model in Eq.~\eqref{eq:glm-1}), $\mathbf{J}(\gamma) = \Vert \gamma \Vert_1 + \frac{1}{2\kappa} \Vert \gamma \Vert_2^2$, the Eq.~\eqref{eq:gsplit-ode} degenerates to \emph{Split Linearized Bregman Inverse Scale Space} (Split LBISS), of which the model selection consistency (the support set of $\gamma^t$ belongs to the true support set when $t\geq t_0$ for some $t_0$) was established in \cite{splitlbi}. In fact, such a statistical analysis in high dimensional space can be traced back to \cite{osher2016sparse} for LBISS with $\nu = 0$ and $D = I$, which is proposed as dynamics form of \emph{Linearized Bregman Iteration} (LBI) that was early known for optimization in compressed sensing and image denoising \cite{osher2005an}. The \cite{huang2018unified} extended such a statistical result to convex loss. The Eq.~\eqref{eq:gsplit-ode} considers more general structural sparsity that incorporates non-negativity under general linear model (GLM), therefore we called it \emph{Generalized Split Linearized Bregman Inverse Scale Space} (GSplit LBISS). 

As a natural extension to generalized linear model and structural sparsity, our GSplit LBISS inhereits the statistical property of LBISS. Notably, the GSplit LBISS will fit signals with structural sparsity first, and then tends to over-fit the random noise that induces the distribution of $y|x$ in Eq.~\eqref{eq:glm-1}. To determine the time $t$ for stopping, we can implement cross-validation, which will be discussed later. At each step in such a regularization solution path, the $\gamma^t$ can be obtained by back-projection of $v^t$ into primal space with structural sparsity as a priori. Specifically, the Eq.~\eqref{eq:gsplit-ode-d} is equivalent to:
\begin{align}
    \label{eq:gsplit-ode-d-equi}
    \gamma^t = \arg\max_{\gamma} \langle \gamma, v^t \rangle - \mathbf{J}(\gamma),
\end{align}
with $\mathbf{J}^{\star}(v^t) := \max_{\gamma} \langle \gamma, v^t \rangle - \mathbf{J}(\gamma)$. We can obtain another $\beta^t_{\mathrm{les}}$, which is the projection of $\beta_{\mathrm{pre}}^t$ onto the subspace of support set of $\gamma^t$, therefore can be regarded the parameters for selecting lesion features. The remainder of such a projection is heavily influenced by procedural bias, which denotes the elements corresponding to large value in such a remainder. The rest with tiny values are regarded as null features/voxels that are not correlated with the disease label, as illustrated by the following orthogonal decomposition of $\beta_{\mathrm{pre}}$: 
\begin{equation}
\label{eq:decompose}
\beta_{\mathrm{pre}} = \mathrm{lesion \ Features} \oplus \mathrm{Procedural \ Bias} \oplus \mathrm{Null \ Features}.
\end{equation}
Such a procedural bias can be learned from data in a supervised way, the extent of which is determined by the hyper-parameter $\nu$ and $t$ (which will be discussed later). Equipped with ability of selecting lesion features guided by $\gamma^t$ to select lesion features and additional capture of procedural bias, the $\beta_{\mathrm{pre}}$ can achieve better prediction performance, as will be shown in our experimental section.  

\begin{remark}
 Note that the $\mathbf{J}(\gamma) = \Vert \gamma \Vert_1 + \frac{1}{2\kappa} \Vert \gamma \Vert_2^2 + \mathbbm{1}(\gamma_V \geq 0)$ is differentiable and smooth. Therefore, according to \cite[Proposition 3.1]{mirror-descent}, we have $\gamma^t \in \nabla \mathbf{J}^{\star}(v^t)$, replaced Eq.~\eqref{eq:gsplit-ode-d} with which the our GSplit LBISS in Eq.~\eqref{eq:gsplit-ode} is equivalent to the dynamics form of mirror descent algorithm (MDA) \cite{nemirovsky1983problem, krichene2015accelerated}. However, the MDA requires that the penalty function $\mathbf{J}$ is smooth, under which the \cite{mirror-descent, krichene2015accelerated} gave the convergent analysis. With existence of noise, the final point (convergent solution $\lim_{t\to \infty} (\beta^t_{\mathrm{pre}}, \gamma^t)$ of LBISS with non-smooth $\mathbf{J}$ will over-fit \cite{osher2016sparse}. Therefore, instead of targeting on the convergent result of Eq.~\eqref{eq:gsplit-ode}, we exploits the 
inverse scale space property to equip GSplit LBISS with early-stopping mechanism, in order to select the solution that satisfies model selection consistency in the whole regularization path. Beyond that, possessed with variable splitting scheme $\frac{1}{2\nu} \Vert D\beta_{\mathrm{pre}} - \gamma \Vert_2^2$, at each time-step $t$ the GSplit LBISS will return a dual parameters: the $\beta^t_{\mathrm{les}}$ with structural sparsity and the dense parameter $\beta_{\mathrm{pre}}$ that can further capture the procedural bias which is highly correlated with $y$, endowed with orthogonal decomposition (Eq.~\eqref{eq:decompose}). 
 \end{remark}

%The capture of procedural bias is affected by hyper-parameter $\nu$ and $t$. With larger value of $\nu$, the $\beta_{\mathrm{pre}}$ has more degree of freedom to capture procedural bias, as illustrated in Fig.~\ref{figure:path}. With smaller value of $\nu$, the parameter space is located into a more lower dimensional space, which may result in $\beta_{\mathrm{les}}$ with more stability, which is important in medical image analysis, as illustrated in Fig.~\ref{figure:illustrate}.

\subsection{Generalized Split Linearized Bregman Iteration: the simple discretization of GSplit LBISS}
\label{sec:gsplit_lbi}

The Eq.~\eqref{eq:gsplit-ode} enjoys a simple discretization, which is called \emph{Generalized Split Linearized Bregman Iteration} (GSplit LBI) in this paper: 
\begin{subequations}
\label{eq:gsplit-lbi}
\begin{align}
\beta^{t+1}_0 & = \beta^{t}_0 - \alpha \kappa \nabla_{\beta_{0}^t} \mathcal{L}(\beta^t_{\mathrm{pre}},\beta^t_0, \gamma^t), \label{eq:gsplit-lbi-a} \\
\beta^{t+1}_{\mathrm{pre}} & = \beta^t_{\mathrm{pre}} - \alpha \kappa  \nabla_{\beta_{\mathrm{pre}}} \mathcal{L}(\beta^t_{\mathrm{pre}},\beta^t_0, \gamma^t), \label{eq:gsplit-lbi-b} \\
v^{t+1} & = v^t - \alpha \nabla_{\gamma} \mathcal{L}(\beta^t_{\mathrm{pre}},\beta^t_0, \gamma^t), \label{eq:gsplit-lbi-c} \\
\gamma^{t+1} & = \kappa \cdot \mathrm{prox}_{\mathbf{J}}(v^{t+1}), \label{eq:gsplit-lbi-d} \\
S^{t+1} & = \mathrm{supp}(\gamma^{t+1}) \label{eq:gsplit-lbi-e} \\
\beta^{t+1}_{\mathrm{les}} & = \mathbf{P}_{S^{t+1},\geq 0}(\beta_{\mathrm{pre}}^{t+1}), \label{eq:gsplit-lbi-f}
\end{align}
\end{subequations}
with $\beta^0 = 0,\beta_{\mathrm{les}}^0 = \beta_{\mathrm{pre}}^0 = 0^{p}, \gamma^0 = v^0 = 0^{p+m}$, and 
\begin{align}
    \mathrm{prox}_{\mathbf{J}}(v) & :=  \arg\min_{\gamma} \frac{1}{2} \Vert \gamma - v \Vert_2^2 + \mathbf{J}(\gamma), \label{eq:prox} \\
    \mathbf{P}_{S,\geq 0}(\beta_{\mathrm{pre}}) & = \arg\min_{\substack{D_{S^c}\beta_{\mathrm{les}} = 0 \\ \beta_{\mathrm{les}} \geq 0}} \Vert \beta_{\mathrm{les}} - \beta_{\mathrm{pre}} \Vert_{2} \label{eq:gamma}
\end{align}
with $S:= \mathrm{supp}(\gamma)$ and it returns the parameter $\beta_{\mathrm{les}}$ for selecting lesion features. With Eq~\eqref{eq:prox}, the Eq.~\eqref{eq:gsplit-lbi-d} is equivalent to $\gamma_V = \kappa \cdot \max(z_V - 1, 0)$ and $\gamma_G = \kappa \cdot \sign(z_G) \cdot \max(|z_G| -1,0)$. For Eq.~\eqref{eq:prox}, since $\mathrm{supp}(\gamma) = \mathrm{supp}(\gamma_V) \cup \mathrm{supp}(\gamma_G)$ and $\mathrm{supp}(\gamma_G) \subseteq E$ defines an edge set on node $V := \{1,...,p\}$, we can find the set of connected components $\mathcal{C}$ of $\tilde{G} = (V,\mathrm{supp}(\gamma_G))$ via \emph{Depth-First Search} (DFS) algorithm as introduced in algorithm~\ref{alg:dfs} with computational complexity $\mathcal{O}(p+m)$. The elements belonging to the same connected component share the same value in terms of $\beta_{\mathrm{les}}$. Suppose there are $T$ connected components for $\tilde{G} = (V,\mathrm{supp}(\gamma_G))$, i.e., $\mathcal{C} = \{i_1,...,i_{n_t}\}_{t=1}^T$, then for all $t \in \{1,...,T\}$ we have
\begin{align}
\beta_{\mathrm{les}}(i_t) =
\begin{cases}
     0, \ \forall i_t \in \mathrm{supp}(\gamma_V) \\
     \max \left(\frac{1}{n_t} \sum_{k=1}^{k=n_t} \beta_{\mathrm{pre}}(i_{k}), 0 \right), \ \forall i_t \in \mathrm{supp}(\gamma_V)^c 
     \end{cases}.
\end{align}

\begin{algorithm}[t]
\caption{DFS to find connected components for graph $G = (V,E)$}
\label{alg:dfs}
	\begin{algorithmic}[1]
	\State Initialize all vertices as not visited and connected components $C = \{\}$.
	\State $\forall v \in V$, 
	\State \quad if $v$ is not visited before, define $C_{\mathrm{new}} = \{ \}$ and call DFSConnect($\mathcal{C}_{\mathrm{new}}$, $v$) 
	\State \quad else, $\mathcal{C} = \mathcal{C} \cup \mathcal{C}_{\mathrm{new}}$.
	\State Output: $\mathcal{C}$.
    \State ----------------------------------------------------------------------
    \Procedure{DFSConnect($\mathcal{C}_{\mathrm{new}}$, $v$)}{$a$}
    \State Input: $\mathcal{C}_{\mathrm{new}}$
    \State Mark $v$ as visited. 
    \State $\mathcal{C}_{\mathrm{new}} = \mathcal{C}_{\mathrm{new}} \cup \{v\}$
    \State For every $u \in N(v) := \{(u,v) \in E\}$ that is not visited, apply DFSConnect($\mathcal{C}_{\mathrm{new}}$, $v$). 
\EndProcedure
	\end{algorithmic}
\end{algorithm}

\begin{figure*}[!h]
\centering
    \includegraphics[width= 2\columnwidth]{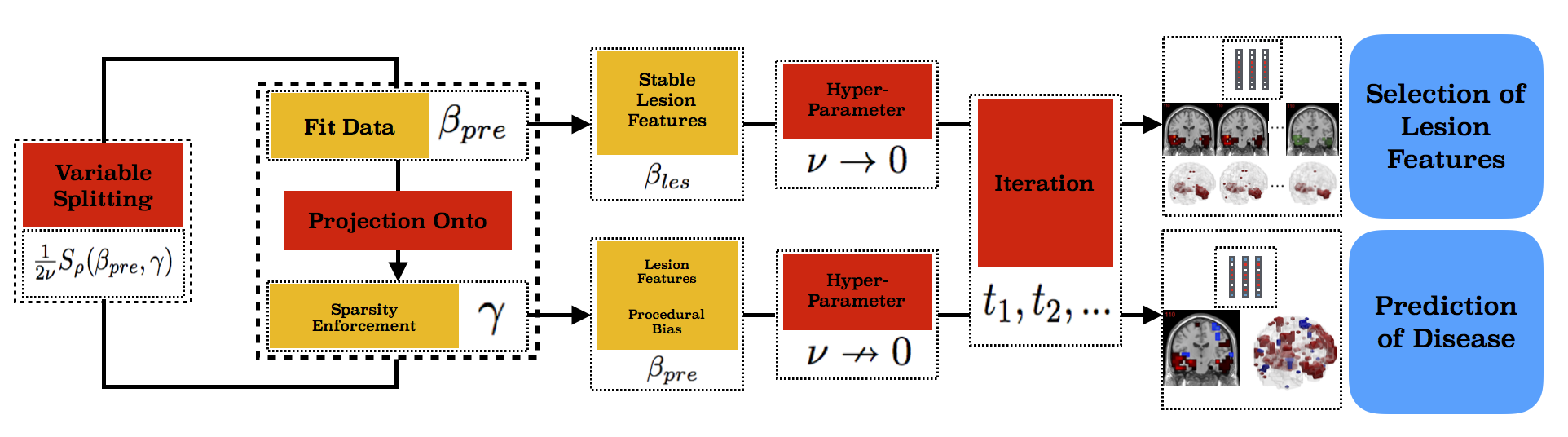}
\caption{Illustration of how \emph{GSplit LBI} works. The gap between $\beta_{\mathrm{pre}}$ for fitting data and $\gamma$ for sparsity is controlled by $\mathbf{S}_{\nu}(\beta_{\mathrm{pre}},\gamma)$. The estimate $\beta_{\mathrm{les}}$, as a projection of $\beta_{\mathrm{pre}}$ on support set of $\gamma$, can be used for stable lesion feature analysis when $\nu \to 0$ (Section~\ref{section-lesion}). When $\nu \nrightarrow 0$ (Section~\ref{section-results}) with appropriately large value, $\beta_{\mathrm{pre}}$ can be used for prediction by capturing both lesion features and procedural bias.}
\label{figure:illustrate}
\end{figure*}

\textbf{Setting hyper-parameters} (i) The $t^k = k \cdot \alpha$ is regularization hyper-parameter, which is often chosen by cross validation \cite{LBI}. Parameter $\rho$ is a tradeoff between geometric clustering and voxel sparsity. Heuristically, Its choice depends on the quality of data, e.g. for data with lower resolution of MRI images, comparably larger $\rho$ is suggested since in this case more clustered features will be selected to suppress the noise. (ii) The $\kappa$ is a damping factor which can be used to reduce bias with a large value. Specifically, it's shown in \cite{LBI} that as $\kappa$ becomes larger, the solution path of LBI looks similar to that of Bregman ISS (Inverse Scale Space), which returns a solution path with bias-free estimators. However, too large $\kappa$ results in more variance and also lower the convergence speed. (iii) The $\alpha$ in LBI for discretization is step size, which is the tradeoffs between statistical and computational issues \cite{libra}. One can generate a `coarse' regularization path with larger value of $\alpha$ though, it fails to satisfy statistical properties, such as recovering true signal set. On the other hand, smaller $\alpha$ makes the solution path `denser', hence lower the computational speed. To ensure the stability of iterations, one may choose $\alpha < \nu / \kappa(1 + \nu \Lambda_{H} + \Lambda_{D}^{2})$ \cite{HUANG2018} with $\Lambda_{A}$ denoted as the largest singular value of general matrix $A$ and $H$ here denotes Hessian matrix of $\mathcal{L}(\beta_0, \beta_{\mathrm{pre}})$ (iv) The choice of $\nu$ is task-dependent, as shown in Fig.~\ref{figure:illustrate}. For lesion feature analysis, $\beta_{\mathrm{les}}$ with a small value of $\nu$ is helpful to enhance stability of feature selection. Note that when $\nu \to 0$, $\Vert \beta_{\mathrm{pre}} - \beta_{\mathrm{les}}\Vert_{2} \to 0$ in the whole path. In this case $\beta_{\mathrm{les}}$ will select lesion features with a good fitness of data. For prediction of disease, $\beta_{\mathrm{pre}}$ with appropriately larger value of $\nu$ has more degree of freedom to leverage procedural bias into prediction. We will discuss it for details later.

\begin{figure*}[!h]
\centering
\begin{minipage}{0.32\linewidth}
	\includegraphics[width= \columnwidth]{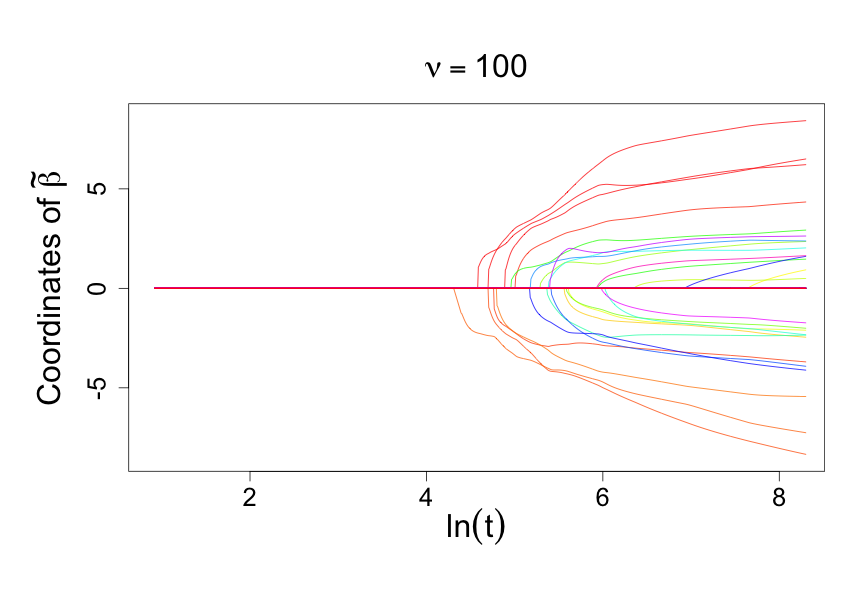}
\end{minipage}
\begin{minipage}{0.32\linewidth}
	\includegraphics[width= \columnwidth]{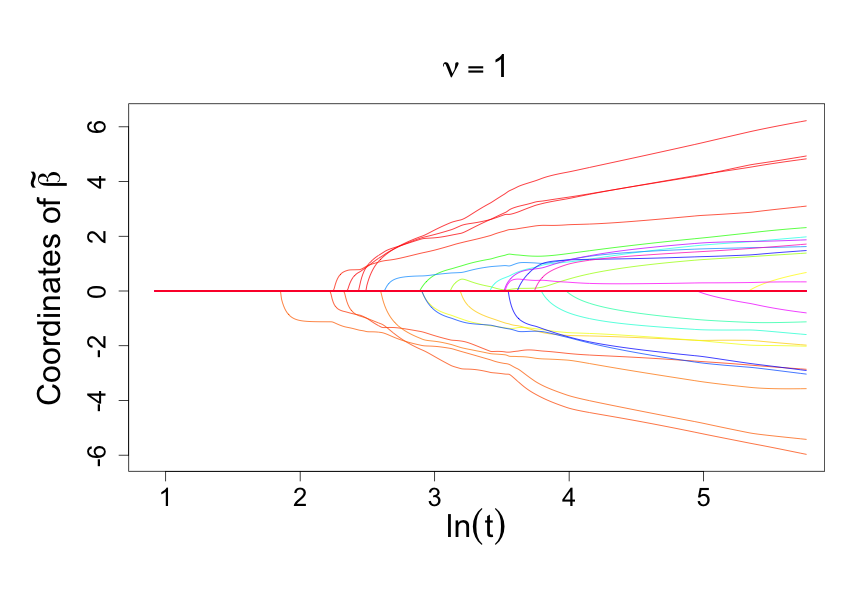}
\end{minipage}
\begin{minipage}{0.32\linewidth}
	\includegraphics[width= \columnwidth]{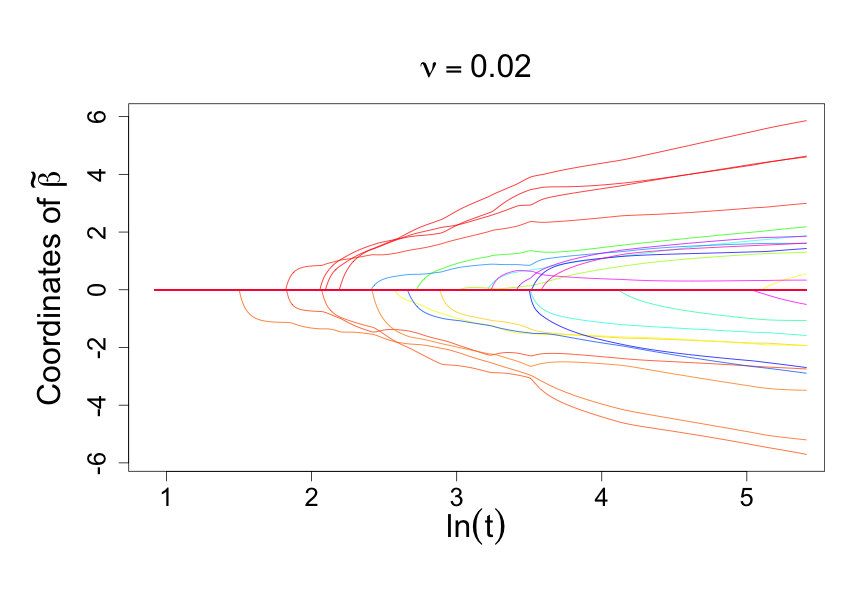}
\end{minipage}
\begin{minipage}{0.32\linewidth}
	\includegraphics[width= \columnwidth]{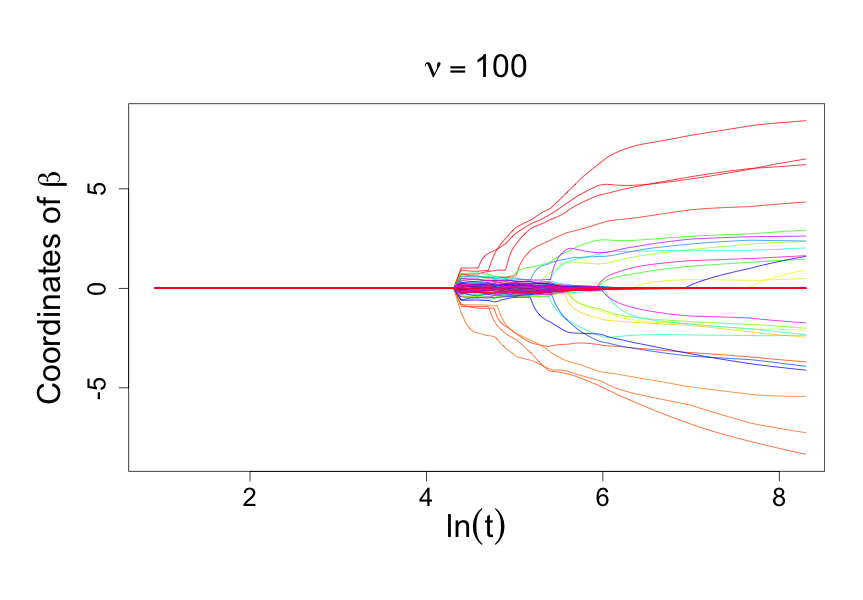}
\end{minipage}
\begin{minipage}{0.32\linewidth}
	\includegraphics[width= \columnwidth]{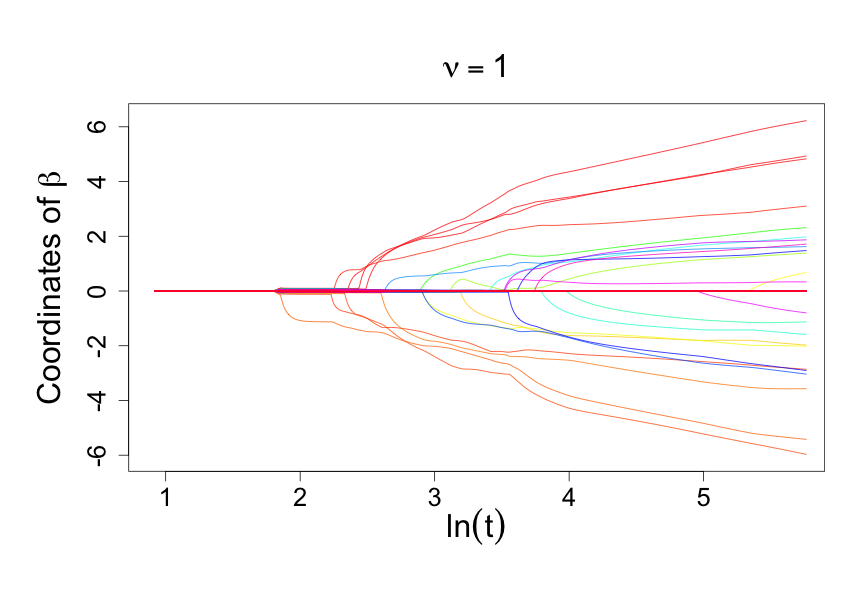}
\end{minipage}
\begin{minipage}{0.32\linewidth}
	\includegraphics[width= \columnwidth]{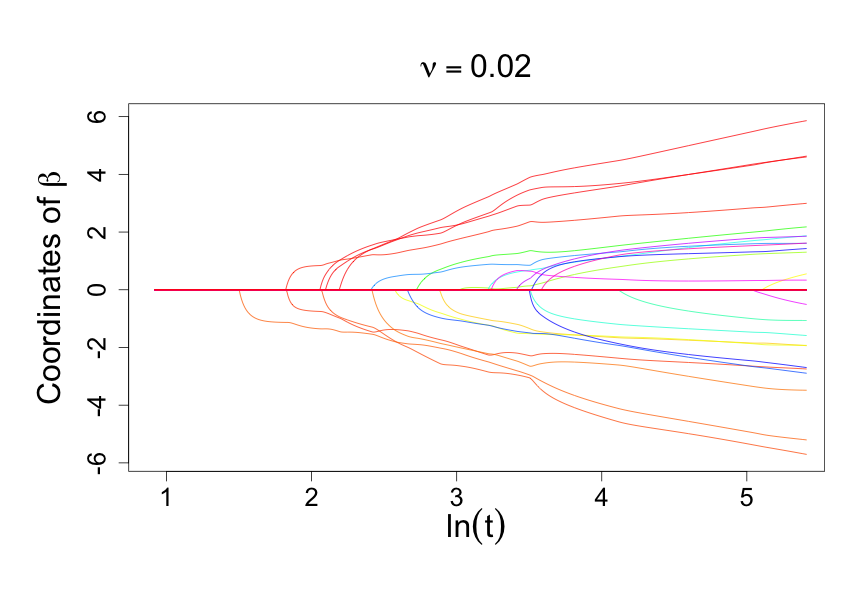}
\end{minipage}
\caption{Comparison of regularized solution path between $\beta_{\mathrm{pre}}$ (denoted as $\beta$ in the figure) and $\beta_{\mathrm{les}}$ (denoted as $\tilde{\beta}$ in this figure) when $\nu = 100,1,0.02$. Each color represents the solution path for each variable. As shown the paths of $\beta_{\mathrm{pre}}$ and $\beta_{\mathrm{les}}$ look more similar with each other as $\nu$ decreases.}
\label{figure:path}
\end{figure*}

\textbf{Exploitation of Procedural Bias} The capture of procedural bias is affected by hyper-parameter $\nu$ and $t$. With larger value of $\nu$, the $\beta_{\mathrm{pre}}$ has more degree of freedom to capture procedural bias, as illustrated in Fig.~\ref{figure:path}. With smaller value of $\nu$, the parameter space of $(\beta_{\mathrm{pre}},\gamma)$ is located into a lower dimensional space, which may result in $\beta_{\mathrm{les}}$ with more stability, as illustrated in Fig.~\ref{figure:illustrate}.

\begin{figure*}[!h]
\centering
\begin{minipage}{0.32\linewidth}
	\includegraphics[width= \columnwidth]{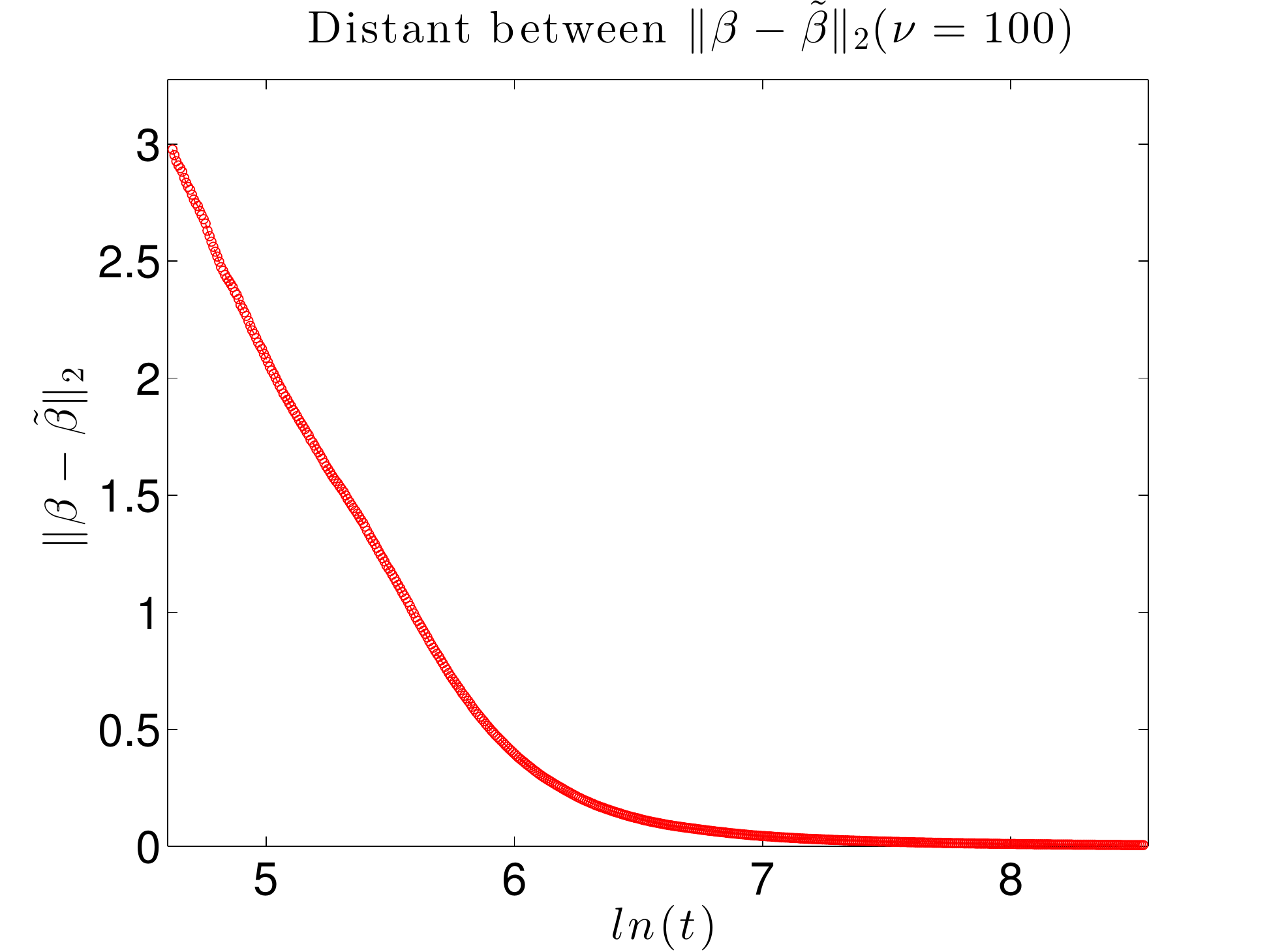}
\end{minipage}
\begin{minipage}{0.32\linewidth}
	\includegraphics[width= \columnwidth]{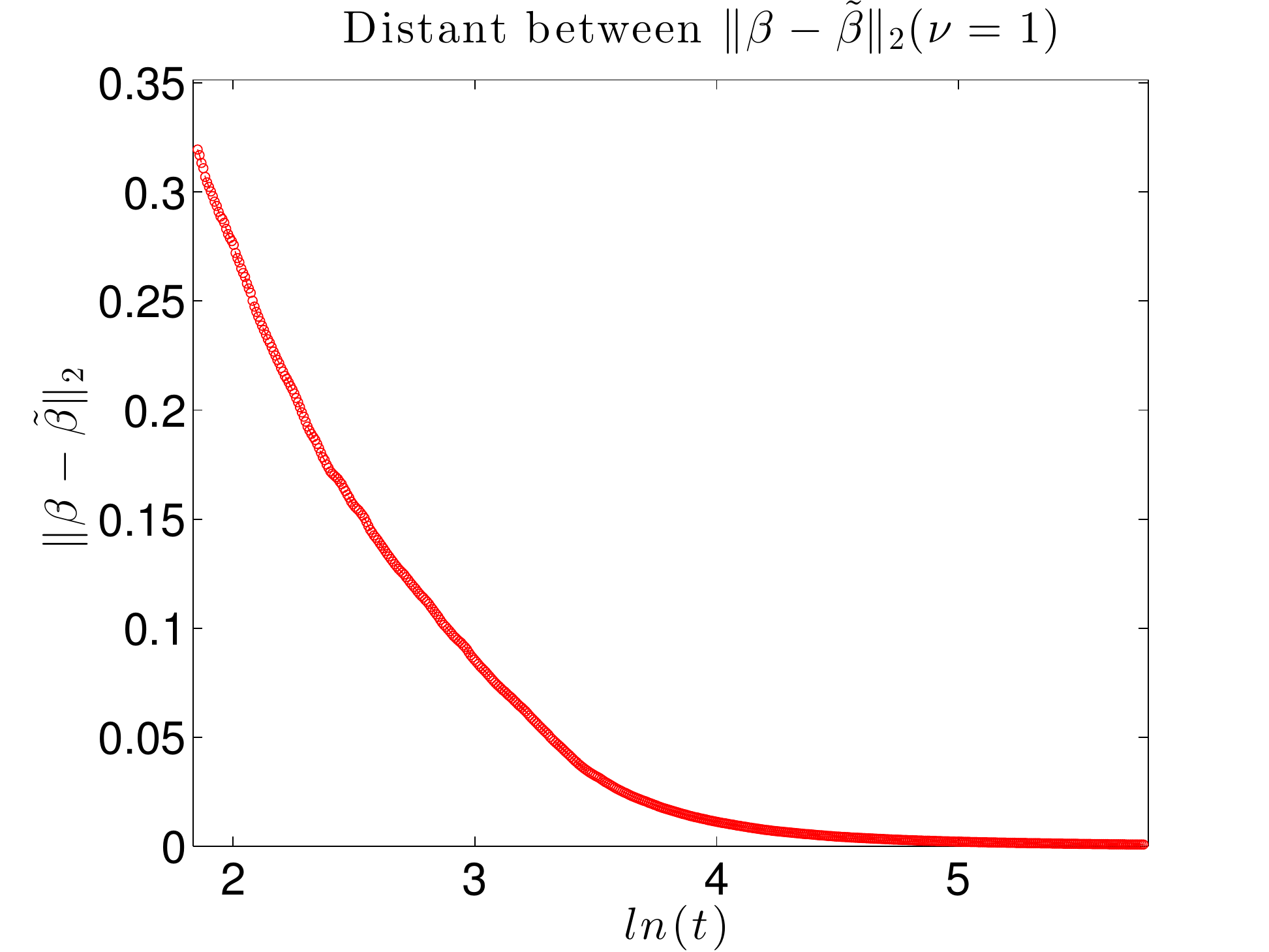}
\end{minipage}
\begin{minipage}{0.32\linewidth}
	\includegraphics[width= \columnwidth]{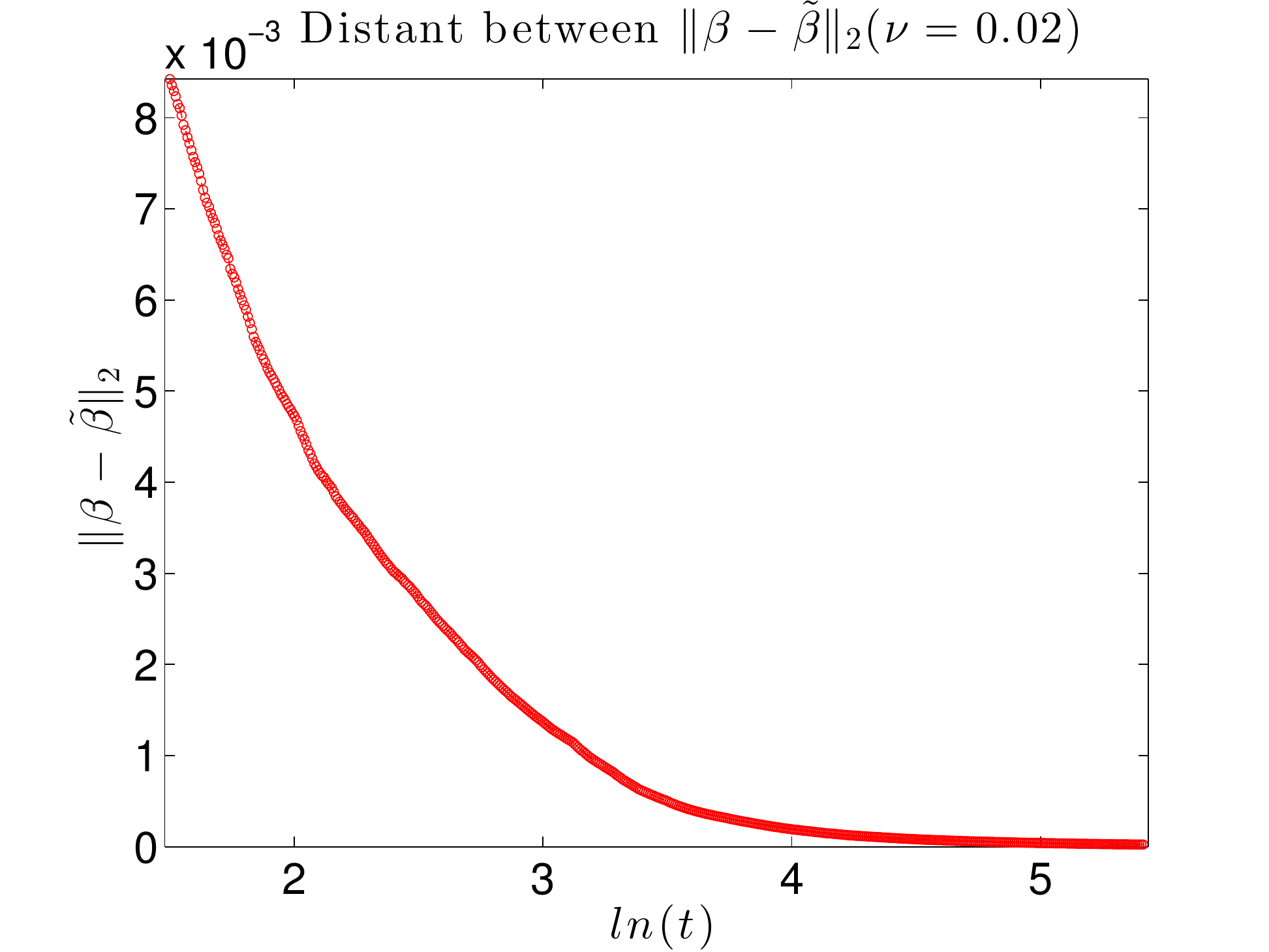}
\end{minipage}
\caption{$\Vert \beta_{\mathrm{pre}} - \beta_{\mathrm{les}} \Vert_{2}$ in the regularized solution path when $\nu = 100,1,0.02$. As $\nu$ decreases, the distance of $\beta_{\mathrm{pre}}(t)$ and $\beta_{\mathrm{les}}(t)$ are tended to be with smaller distance.Her "$\beta$" denotes $\beta_{\mathrm{pre}}$ and "$\tilde{\beta}$" denotes $\beta_{\mathrm{les}}$.}
\label{figure:curve}
\end{figure*}

As $t$ grows, the $\beta_{\mathrm{les}}(t)$ is with less sparse regularization effect, hence converges to $\beta_{\mathrm{pre}}$, i.e.,  $\Vert \beta_{\mathrm{pre}}(t) - \beta_{\mathrm{les}}(t) \Vert_{2} \to 0$, as illustrated in Fig.~\ref{figure:curve}. In this case (i.e. $t$ is large), $\beta_{\mathrm{les}}(t)$ may overfit to the noise, hence can select redundant features, which will also deterioriate the performance of $\beta_{\mathrm{pre}}$. Hence, an early stopping strategy is necessary to implement to avoid overfitting, as mentioned earlier. 

\begin{comment}
%In The dense estimator $\beta_{\mathrm{pre}}$ can be Following the Linearized Bregman Iteration \cite{LBI}, $\beta_{\mathrm{pre}}$ and $\beta_{\mathrm{les}}$ will be more similar on features selected by $\beta_{\mathrm{les}}$. In more detail, note that when $t = 0$, $\beta_{\mathrm{les}}(t) = 0$ and $\beta_{\mathrm{pre}}(t)$ is the graph laplacian regularizer with penalty factor $\frac{1}{2\nu}$. As $t$ progresses, the gap between $\beta_{\mathrm{pre}}(t)$ and $\beta_{\mathrm{les}}(t)$ will decrease in terms of $\Vert \beta_{\mathrm{pre}}(t) - \beta_{\mathrm{les}}(t) \Vert_{2}$ for every $\nu$, as shown in Fig.~\ref{figure:curve}.
%The difference between $D\beta_{\mathrm{les}}(t)$ and $\gamma(t)$ will be lower in the whole path for smaller value of $\nu$, as shown in Fig.~\ref{figure:curve} and~\ref{figure:path}. This implies that the vector of $(\beta_{\mathrm{pre}},\gamma)$ are located in a lower dimensional space. The implication is 2-folds: on one hand the capability of $\beta_{\mathrm{pre}}$ to capture other relevant features for better fitness of data is lowered; while on the other hand, the $\beta_{\mathrm{les}}$ can select features with more stability.
\end{comment}

\textbf{Computational complexity} For $\ell_{1}$-minimization, for each regularization hyper-parameter $\lambda$, a popular iterative algorithm to solve the minimization problem is the well-known iterative soft-thresholding algorithm (ISTA) (\cite{fista} and references therein). To tune $\lambda$, one needs to run ISTA until convergence for each $\lambda$, which is time-consuming. In a contrast, LBI returns the full regularization path where estimators in each iteration is function of sparsity level \cite{LBI,LBI-logistic} regularized by $t_{k}$. The model selection consistency of such a regularization path has been theoretically discussed in \cite{LBI} for the square loss and in \cite{glbi} for general losses (including the logistic regression). More importantly, at each $t_{k}$ the minimizer of sub-problem and the $\beta_{\mathrm{les}}$ can be written in a closed form achieved by soft-thresholding operator and DFS algorithm. In this respect, LBI only runs a single path with early stopping time $t^*$ determined via cross-validation. In addition, by taking step size $\alpha$ small enough and $\kappa$ large enough to ensure the stability of iterations, one can achieve sign-consistency and minimax optimal $\ell_{2}$-error bounds \cite{LBI} under the same conditions as $\ell_1$-regularization (Lasso) but with less bias \cite{osher2016sparse}; while $\ell_{1}$-minimization deteriorates true signals due to its bias. For each iteration, the time complexity of LBI and ISTA with logistic regression loss is $O(np)$; LBI returns the full regularization path at $O(npk)$ where $k$ is the number of iterations, while with $O(npk)$ iterations ISTA only returns a single estimate at particular $\lambda$. For GSplit LBI, with cost at computation for gradient of variable splitting term, the time complexity is $O(npk + mpk)$. Although with additional $O(mpk)$, it can cost much less than $\ell_{1}$-minimization which costs $O(n_{\lambda}npk)$, where $n_{\lambda}$ is the number of $\lambda$ in the grid of regularization parameters. However, it should be noted that for GSplit LBI, the number of iterations, $k$ is determined by $\alpha$, $\kappa$ and $\nu$. 
%We leave the detailed discussion to section~\ref{setting-parameters}.

\begin{figure}[!h]
\begin{center}
\begin{minipage}[t]{0.45\linewidth}
    \includegraphics[width= \columnwidth]{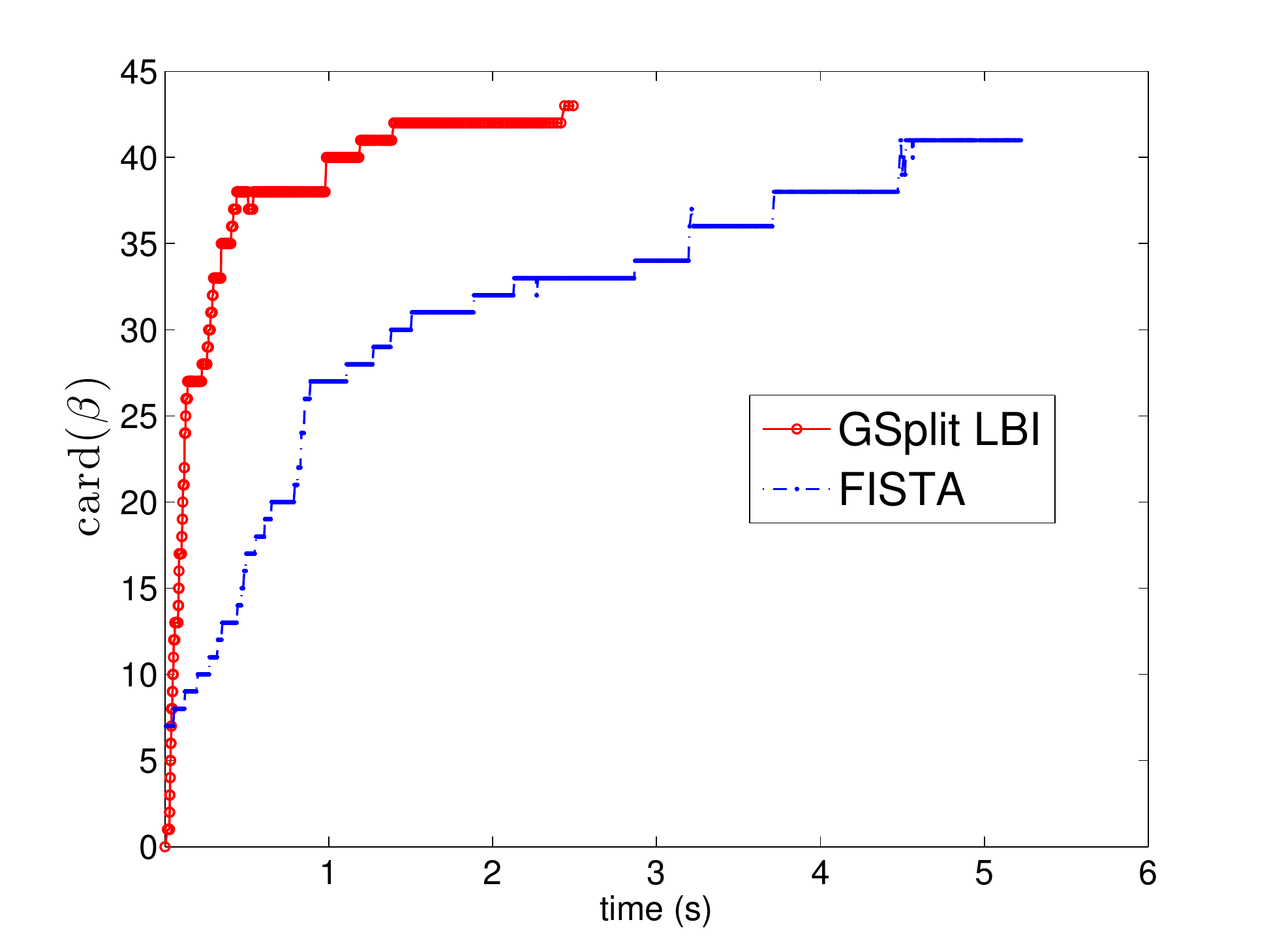}{\begin{center} (a) \end{center}}
\end{minipage}
\begin{minipage}[t]{0.45\linewidth}
    \includegraphics[width= \columnwidth]{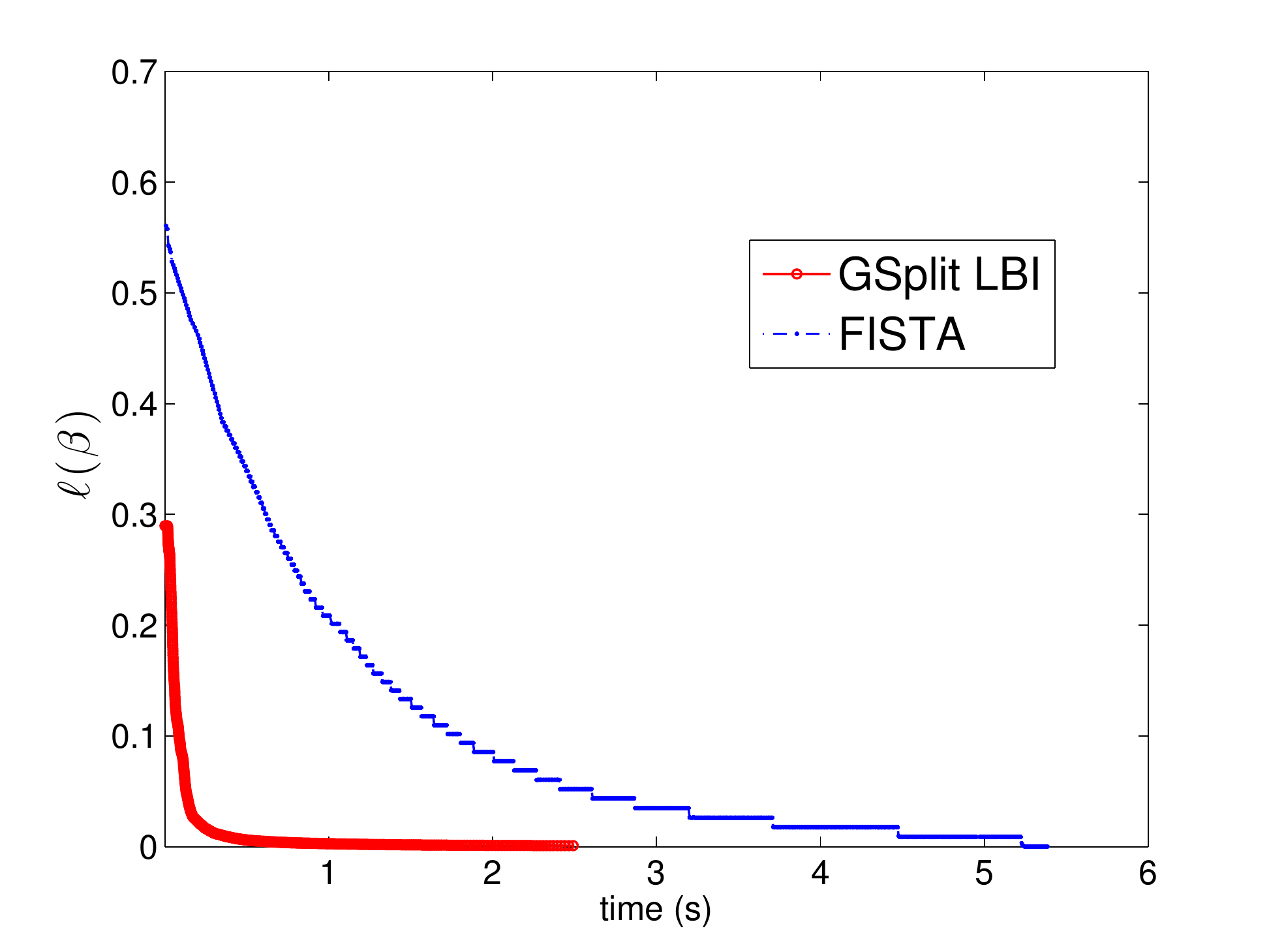}{\begin{center} (b) \end{center}}
\end{minipage}	
\begin{minipage}[t]{0.45\linewidth}
    \includegraphics[width= \columnwidth]{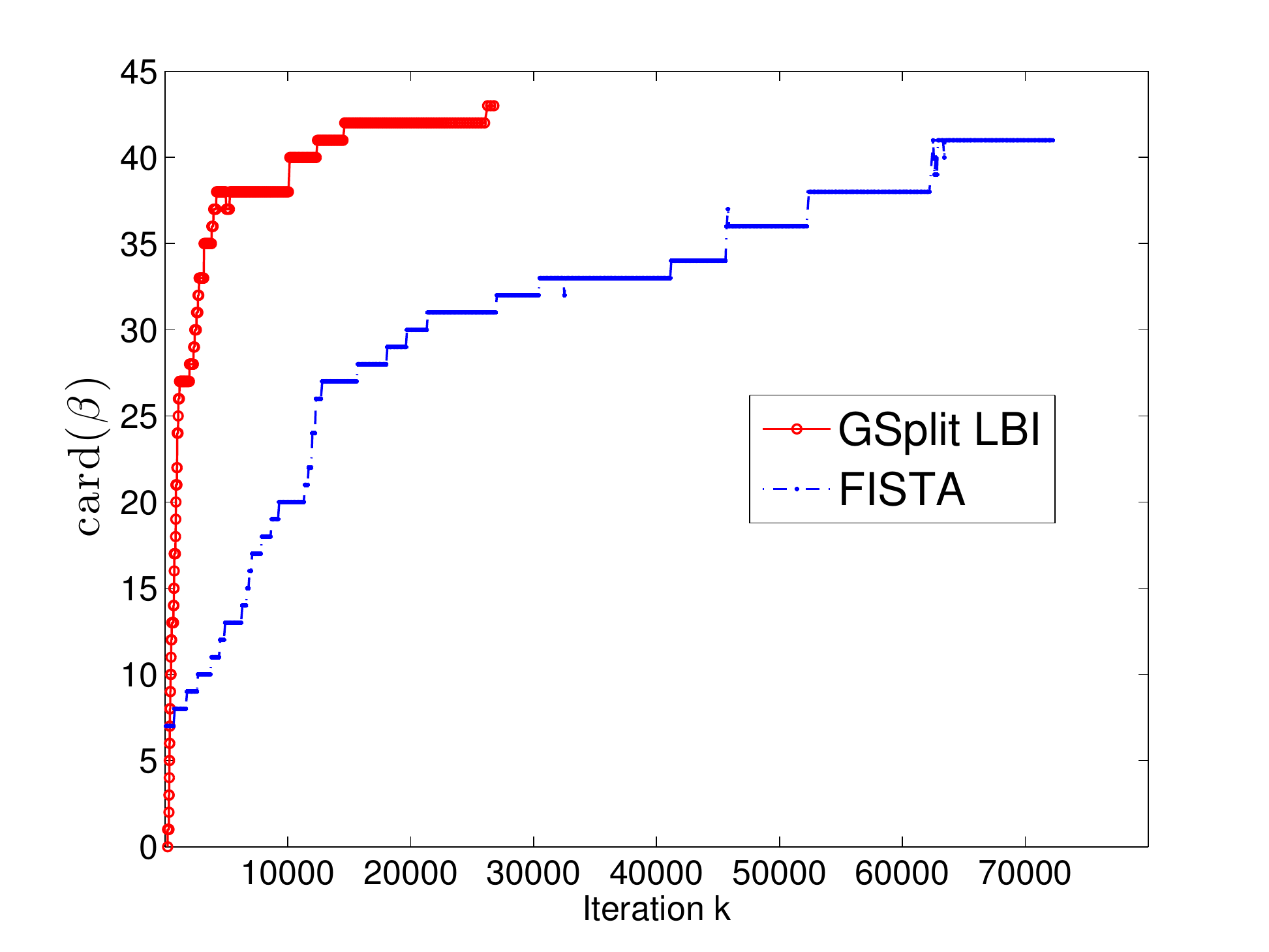}{\begin{center} (c) \end{center}}
\end{minipage}
\begin{minipage}[t]{0.45\linewidth}
    \includegraphics[width= \columnwidth]{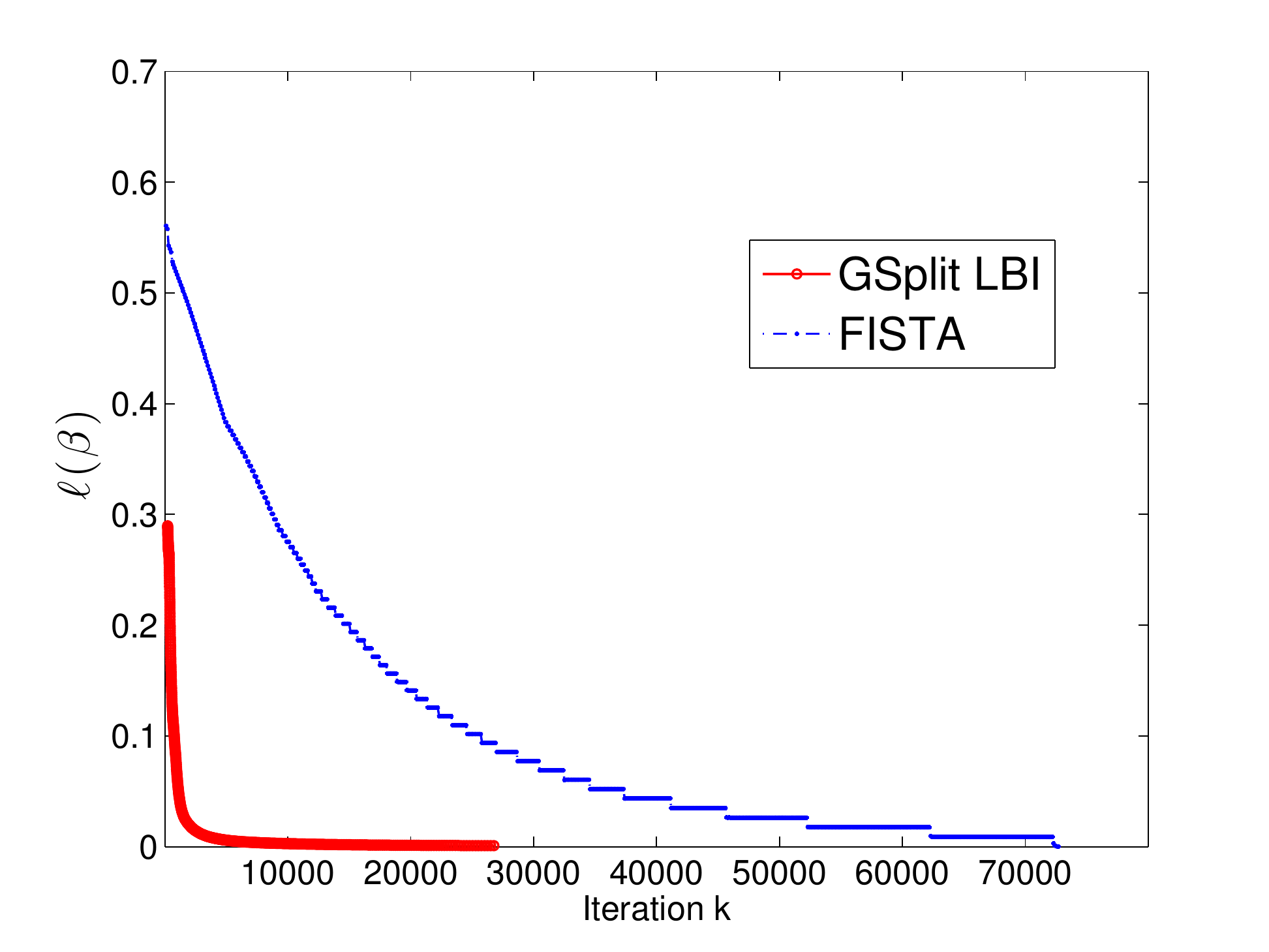}{\begin{center} (d) \end{center}}
\end{minipage}	
\end{center}
\caption{(a)$\&$(b) show the function curve of support cardinality and loss versus computational time. (c)$\&$(d) show the them versus number of iterations.}
\label{figure:compare-fista-lbi}
\end{figure}

A simulation experiment is conducted to compare computational efficiency of GSplit LBI and $\ell_{1}$-minimization, in terms of $card(\beta_{\mathrm{les}}) := | \mathrm{supp}(\beta_{\mathrm{les}}) |$ and $\ell(\beta_{\mathrm{pre}}^k)$. For experimental setting, we set $n = 100$ and $p = 80$, $D = I$ and $X \in \mathbb{R}^{n \times p}$ and $X_{i,j} \sim N(0,1)$. $\beta^{\star}_{i} =  2$ for $1 \leq i \leq 4$, $\beta^{\star}_{i} =  -2$ for $5 \leq i \leq 8$ and 0 otherwise, $y$ is generated via logit model given $X$ and $\beta^{\star}$. For GSplit LBI, we run a single path and compute $card(\beta_{\mathrm{les}}^{k})$ and $\ell(\beta_{\mathrm{pre}}^k)$ at each iteration. For $\ell_{1}$-minimization, we set a sequence of $\lambda$ and run FISTA \cite{fista}, an accelerate version of ISTA for every $\lambda$. Besides, we employ the computation strategy, i.e. take the solution of $\lambda_{k}$ as the initial solution for the next $\lambda_{k+1}$. For each $\lambda$, the algorithm stops as long as either of following stopping criteria is satisfied:
\begin{equation*}
\Vert \beta^{k+1} - \beta^{k} \Vert_{1} / p < 10^{-8}, \ \Vert \nabla \ell(\beta^{k}) \Vert_{\infty} < 10^{-6}
\end{equation*}
%For simplicity, we adopt the same setting in section~\ref{sec:gsplit_lbi}. 
All experiments were implemented in MATLAB R2014a 64-bit on a MacBook Pro laptop with Mac OSX 10.12.4, 2.6 GHz Intel Core i5 and 8 GB 1600 MHz DDR3 memory. Fig.~\ref{figure:compare-fista-lbi} shows the function curve of cardinality of solution support and loss versus the accumulated computational time and number of iterations. As shown, the GSplit LBI is convergent faster than FISTA in terms of cardinality of support and loss function. 

\section{Simulation}
\label{sec:simulation}

\begin{figure}
\centering
\includegraphics[width= 0.9\columnwidth]{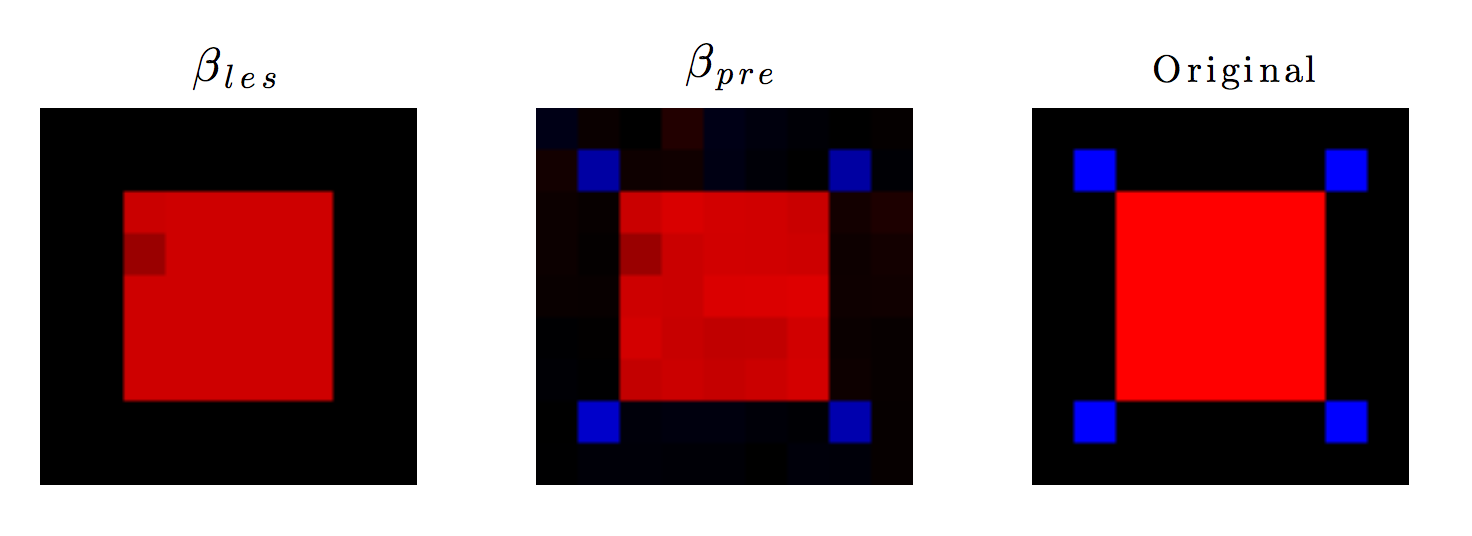}
\caption{Simulation results. The lesion features(red) can be recovered by both $\beta_{\mathrm{les}}$ and $\beta_{\mathrm{pre}}$; the procedural bias(blue) can be recovered by $\beta_{\mathrm{pre}}$}
\label{figure:sim}
\end{figure}

In this experiment, we conduct a simulation experiment to illustrate the effect of selecting both lesion features and procedural bias for GSplit LBI. To simulate voxel-based neuroimage scenario, the lesion features are set to be sparse, positively correlated with disease label and clustered into a region (Red Center in ``original" figure in Fig.~\ref{figure:sim}); while the ground truth of procedural bias (Blue Corner in ``original" figure in Fig.~\ref{figure:sim}) is less clustered and negatively correlated with disease label. We show in the following that (1) $\beta_{\mathrm{les}}$ capture lesion features (2) $\beta_{\mathrm{pre}}$ capture both kinds of features. 

In details, we set $N = 400$. The feature space is a 2-d squared images with $p = 9\times 9= 81$ pixels. We denote . For the true signal vector $\beta^{\star} \in \mathbb{R}^{9\times 9}$, the center with size $5 \times 5$ representing lesion features are assigned with value 3; while the four corner pixels representing procedural bias are with value -3; other pixels are assigned with value 0. $X \in \mathbb{R}^{N \times p}$ denotes $N$ i.i.d samples with each generated from $N(0,I_{p})$, i,e.,
\begin{align}
P(y_{i} = \pm 1) = \frac{\mathrm{exp}(\left<x_{i}, \beta^{\star}\right> \cdot y_{i})}{1 + \mathrm{exp}(\left<x_{i}, \beta^{\star}\right> \cdot y_{i})} \nonumber
\end{align}
where $x_{i}$ denotes the $i$th row/sample of $X$. For GSplit LBI, we set $\kappa = 80$, $\nu = 2$ and $\alpha = \nu/ \kappa(1 + \nu \Lambda^{2}_{X} + \nu \Lambda^{2}_{D})$. 

After getting $\beta_{\mathrm{les}}^t$ and $\beta_{\mathrm{pre}}^t$ for some $t$, the positive and negative coefficients can be projected to \textbf{Red} and \textbf{Blue} color, respectively. In more details, for positive features, their \textbf{Red} channel value is normalized by dividing the greatest value. Besides, the \textbf{Green} and \textbf{Blue} channels value of positive features are all kept to 0. The procedure is similar to the projection of the negative features into \textbf{Blue} color. After projection, It can be shown form Fig.~\ref{figure:sim} that $\beta_{\mathrm{les}}$ can successfully recover lesion features (Red Center), the $\beta_{\mathrm{pre}}$ can successfully recover both lesion features and procedural bias (Blue Corner).

\section{Application to ADNI dataset}
\label{sec:adni}
In this section, we apply our method on real-world data of Alzheimer's Disease. 

\subsection{Acquisition of data}
The data are obtained from Alzheimer’s Disease Neuroimaging Initiative (ADNI) \url{http://www.loni.ucla.edu/ADNI} database. Starting from October 2004, ADNI is a long-period non-profit project that was originally funded by National Institute on Aging (NIA), National Institute of Bioimaging and Bioengineering (NIBIB), National Institutes of Health (NIH), pharmaceutical Industry and other private foundations. ADNI GO was launched in 2010 to continue the study of ADNI and started to focus on participants who exhibit the very beginning stages of memory loss. In 2011, ADNI 2 was launched as the third phase of ADNI study. One of the major goal is to efficiently predict mild cognitive impairment (MCI) and early Alzheimer's disease(AD) in clinical practice \cite{ADNI}. 

The ADNI contains various types of data, including images from sMRI, PET and CSF biomarkers. In this paper, we implement analysis on sMRI, which shows high rates of brain atrophy for AD and hence has high statistical power for determining treatment effects \cite{ADNI}. The data are acquired from 1.5T and 3.0T (namely 15 and 30) field strength Magnetic Resonance Imaging scanners. Specifically, the 15 dataset contains 64 AD patients, 90 NC and 110 MCI patients; while 30 dataset contains 66 AD patients, 110 NC and 106 MCI. The overall subject information is listed in table~\ref{table:1.5T-subject} and~\ref{table:3.0T-subject}. \footnote{For subject IDs, please refer to data statement https://github.com/sxwxiaoxiao/ADNI/blob/master/GSplit\text{\%}20LBI\text{\%}20SID.pdf.}

\begin{table}[!h]
\scriptsize
\centering
\caption{The Demographic data for subjects from 1.5T MRI scan. MMSE=Mini-Mental State Examination}
\begin{tabular}{|c|c|c|c|c|c|c|c|c|c|c}
\hline
1.5T & AD & MCI & NC \\
\hline
Size & 64 & 110 & 90  \\
\hline
Gender (M/F) & 43/21  & 77/33 &  61/29  \\
\hline
Age (mean $\pm$ SD) & 74.55 $\pm$ 7.22 & 73.93 $\pm$ 7.56 & 75.07 $\pm$ 4.46  \\
\hline
Education (mean $\pm$ SD) & 14.72 $\pm$ 3.11 & 16.30 $\pm$ 2.84 & 16.49 $\pm$ 2.59  \\
\hline
Hand (R/L) & 61/3 & 99/11 & 83/7  \\
\hline
MMSE & 23.45 $\pm$ 2.00 & 27.28 $\pm$ 1.838 & 28.96 $\pm$ 1.04 \\
\hline
\end{tabular} 
\label{table:1.5T-subject}
\end{table}

\begin{table}[!h]
\scriptsize
\centering
\caption{The Demographic data for subjects from 3.0T MRI scan. MMSE=Mini-Mental State Examination}
\begin{tabular}{|c|c|c|c|c|c|c|c|c|c|c}
\hline
3.0T & AD & MCI & NC \\
\hline
Size & 66 & 106 & 110 \\
\hline
Gender (M/F) & 43/23  & 67/39 & 54/56  \\
\hline
Age (mean $\pm$ SD) &  74.38 $\pm$ 9.32 & 72.25 $\pm$ 7.50 & 73.07 $\pm$ 5.92 \\
\hline
Education (mean $\pm$ SD) & 15.59 $\pm$ 3.01 & 16.36 $\pm$ 2.56 & 16.65 $\pm$ 2.58 \\
\hline
Hand (R/L) & 60/6 & 98/8 & 99/11 \\
\hline
MMSE & 23.11 $\pm$ 2.05 & 27.41 $\pm$ 1.74 & 28.93 $\pm$ 1.26 \\
\hline
\end{tabular} 
\label{table:3.0T-subject}
\end{table}

\subsection{Image Preprocessing}
The image preprocessing follows the DARTEL (Diffeomorphic Anatomical Registration Through Exponentiated Lie Algebra) Voxel-based morphometry (VBM) pipeline \cite{VBM}. VBM is the commonly adopted method to study the of local differences of gray matter concentration among population of subjects. During this process, the preprocessing of registration is necessary, for which the DARTEL is implemented. In the spatial normalization step during registration, the brain templates based on anatomical images of subject groups are made and normalized onto the standard Montreal Neurological Institute template, which makes DARTEL be able to achieve more accurate inter-subject registration of brain images. After implementing this preprocessing step, the final input features consist of 2,527 $8\times 8 \times 8 mm^{3}$ size voxels with average values in GM population template greater than 0.1. Experiments are then designed on four tasks, 1.5T AD/NC, 3.0T AD/NC, 1.5T MCI/NC and 3.0T MCI/NC (namely, 15ADNC, 30ADNC, 15MCINC and 30MCINC, respectively).

\subsection{Validation} 

To evaluate the performance of our model, we used a train-test split strategy to compute the classification accuracy (acc), sensitivity (sen) and specificity (spe), which are defined as
\begin{align}
\mathrm{acc} & = \frac{\# \{ i | \hat{y}_{i} = y_{i} \} }{N} \nonumber \\
 \mathrm{sen} & = \frac{\# \{ i | \hat{y}_{i} = y_{i} = -1 \} }{\# \{ i | y_{i} = -1\} } \nonumber \\
 \mathrm{spe} & = \frac{\# \{ i | \hat{y}_{i} = y_{i} = 1 \} }{\# \{ i | y_{i} = 1\} }\nonumber
\end{align} 
where $\hat{y}$ is the predicted outcome (-1 denotes AD). In more details, we split datasets into $70\%$ training datasets and $30\%$ test datasets. Under exactly the same experimental setup, comparisons are made between GSplit LBI and other classifiers: SVM, MLDA(univariate model via t-test + LDA) \cite{MLDA}, Graphnet \cite{graphnet}, lasso \cite{lasso}, Elastic Net, TV+$L_{1}$ and $n^{2}$GFL. The optimal hyper-parameters of each model are determined via 8-fold cross-validation on training set. Then $\beta_{\mathrm{pre}}$ for prediction on test set is learned on the whole training set implemented with the optimal parameters. To reduce random partition bias, we repeated this procedure for 10 times. For GSplit LBI, the $\kappa$ and $\nu$ are set to 10 and 0.2, respectively; $\alpha = \nu / \kappa(1 + \nu \Lambda_{X}^{2} + \Lambda_{D}^{2})$\footnote{For logit model, $\alpha < \nu / \kappa(1 + \nu \Lambda_{H}^2 + \nu \Lambda_{X}^2)$ since $\Lambda_{X} > \Lambda_{H}$.}. For GSplit LBI, $\rho$ is chosen from $\{0.5,1,..,5\}$, The regularization coefficient $\lambda$ is ranged in $\{0,0.05, 0.1,...,0.95,1,2,5,10\}$ for lasso and $2^{\{-20,-19,...,0,...,20\}}$ for SVM. For other models, parameters are optimized from $\lambda:\thinspace \{0.05,0.1,...,1,2,5,10\}$ and $\rho:  \thinspace \{0.5,1,..,5\}$ (in addition, the mixture parameter $\alpha: \{0,0.05,...,0.95\}$ for Elastic Net).

\section{Results and Analysis}

We firstly compare GSplit LBI with other models according performance on 8-fold cross-validation in training set. It's noted that although K-fold cross-validation may underestimate the prediction error, it can measure the optimistic performance of one's model that it can achieve. As shown from table~\ref{table:cv} that, the results of our model are better or comparable than other models on all tasks. Further, to accurately access the prediction power of GSplit LBI and other models, the results on $30\%$ held-out test sets is reported. Such held-out test sets can measure the generalization to new data, hence the performance on those sets provide the true classification error. To reduce bias of randomness, the process is implemented for 10 times and the mean values are reported. It's shown in table~\ref{table:test} that GSplit LBI also yields better or comparable out-of-sample classification error than other models, with accuracies of $86.31\%$, $88.08\%$, $62.67\%$ and $65.94\%$ on 15ADNC, 30ADNC, 15MCINC and 30MCINC, respectively. Note that Graphnet can yield better results than TV + $\ell_{1}$ and $n^2$GFL since it relaxes the piece-wise constant enforcement. However, it can select more redundant features than necessary, which will be discussed later. 

\begin{table}[!h]
\caption{Comparisons of GSplit LBI with other models(cross-validation)}
\begin{center}
\scriptsize
\begin{tabular}{|p{1.18cm}|p{0.8cm}|p{0.8cm}|p{0.8cm}|p{0.8cm}|p{0.8cm}|p{0.8cm}|p{0.8cm}|p{0.8cm}|p{0.8cm}|p{0.8cm}|p{0.8cm}|p{0.8cm}|p{0.8cm}|p{0.8cm}|p{0.8cm}|p{0.8cm}|p{0.8cm}|}
\hline
 & \multicolumn{3}{|c|}{15ADNC } & \multicolumn{3}{|c|}{30ADNC} \\
\hline
 & ACC & SEN & SPE &  ACC & SEN & SPE  \\
\hline
MLDA &  $84.54\%$ & $83.33\%$ & $85.40\%$ & $87.50\%$ & $90.21\%$ & $85.84\%$ \\
\hline
SVM &  $81.94\%$ & $59.11\%$ & $93.69\%$ & $86.94\%$ & $74.93\%$ & $95.24\%$ \\
\hline
Lasso &  $82.69\%$ & $74.22\%$ & $88.73\%$ & $85.00\%$ & $72.98\%$ & $92.34\%$ \\
\hline           
Elastic Net & $83.52\%$ & $71.78\%$ & $91.90\%$ & $88.95\%$ & $77.02\%$ & $96.23\%$  \\
\hline
Graphnet & $85.09\%$  & $76.00\%$ & $91.27\%$ & $87.26\%$ & $73.19\%$ & $95.84\%$  \\
\hline
TV + $\ell_{1}$ & $83.16\%$ & $74.89\%$ & $88.73\%$ & $85.73\%$ & $74.89\%$ & $92.34\%$ \\
\hline
$n^{2}$GFL & $83.70\%$ & $74.89\%$ &$90.00\%$  &  $84.84\%$ & $69.58\%$ & $94.16\%$ \\
\hline 
Ours($\beta_{\mathrm{pre}}$) & $\textbf{87.22\%}$ & $\textbf{81.33\%}$ & $\textbf{91.43\%}$ & $\textbf{89.11\%}$ & $\textbf{81.49\%}$ & $\textbf{93.77\%}$ \\
\hline
& \multicolumn{3}{|c|}{15MCINC} & \multicolumn{3}{|c|}{30MCINC} \\
\hline
& ACC & SEN & SPE & ACC & SEN & SPE \\
\hline
MLDA &  $66.19\%$ & $63.16\%$ & $69.89\%$ & $67.04\%$ & $65.33\%$ & $68.70\%$ \\
\hline
SVM &  $70.29\%$ & $74.54\%$ & $63.81\%$ & $65.13\%$ & $51.82\%$ & $78.05\%$ \\
\hline
Lasso &  $63.93\%$ & $70.78\%$ & $55.56\%$ & $65.07\%$ & $60.67\%$ & $69.35\%$ \\
\hline
Elastic Net & $72.86\%$ & $78.70\%$ & $65.71\%$ & $70.20\%$ & $66.13\%$ & $74.16\%$ \\
\hline
Graphnet & $72.50\%$  & $78.05\%$ & $65.71\%$ & $69.34\%$  & $60.53\%$ & $77.92\%$ \\
\hline
TV + $\ell_{1}$ & $68.36\%$ & $72.73\%$  & $63.02\%$  & $66.18\%$ & $62.67\%$  & $69.61\%$ \\
\hline
$n^{2}$GFL & $69.14\%$ & $76.62\%$ & $60.00\%$ & $68.23\%$ & $64.00\%$ & $72.34\%$ \\
\hline
Ours($\beta_{\mathrm{pre}}$) &$\textbf{72.93\%}$  & $\textbf{78.31\%}$ & $\textbf{66.35\%}$ &$\textbf{70.72\%}$  & $\textbf{68.67\%}$ & $\textbf{72.65\%}$ \\
\hline
\end{tabular}
\end{center}
\label{table:cv}
\end{table}
%train

%test

\begin{table}[!h]
\caption{Comparisons of GSplit LBI with other models on test dataset}
\begin{center}
\scriptsize
\begin{tabular}{|p{1.18cm}|p{0.8cm}|p{0.8cm}|p{0.8cm}|p{0.8cm}|p{0.8cm}|p{0.8cm}|p{0.8cm}|p{0.8cm}|p{0.8cm}|p{0.8cm}|p{0.8cm}|p{0.8cm}|p{0.8cm}|p{0.8cm}|p{0.8cm}|p{0.8cm}|p{0.8cm}|}
\hline
&  \multicolumn{3}{|c|}{15ADNC } & \multicolumn{3}{|c|}{30ADNC} \\
\hline
 & ACC & SEN & SPE &  ACC & SEN & SPE  \\
 \hline
MLDA &  $84.35\%$ & $81.05\%$ & $86.67\%$ & $86.92\%$ & $90.53\%$ & $84.85\%$ \\
\hline
SVM &  $83.49\%$ & $68.42\%$ & $94.07\%$ & $85.00\%$ & $68.42\%$ & $94.55\%$ \\
\hline
Lasso &  $86.31\%$ & $77.89\%$ & $92.22\%$ & $82.12\%$ & $66.32\%$ & $91.21\%$ \\
\hline           
Elastic Net & $86.31\%$ & $76.32\%$ & $93.33\%$ & $86.15\%$ & $72.11\%$ & $94.24\%$  \\
\hline
Graphnet & $84.57\%$  & $74.21\%$ & $91.85\%$ & $86.92\%$ & $71.58\%$ & $95.76\%$  \\
\hline
TV + $\ell_{1}$ & $79.35\%$ & $68.95\%$ & $86.67\%$ & $84.04\%$ & $68.42\%$  & $93.03\%$  \\
\hline
$n^{2}$GFL & $83.70\%$ & $71.58\%$ &$92.59\%$  &  $83.46\%$ & $67.89\%$ & $92.42\%$ \\
\hline 
Ours($\beta_{\mathrm{pre}}$) & $\textbf{86.31\%}$ & $\textbf{78.95\%}$ & $\textbf{91.48\%}$ & $\textbf{88.08\%}$ & $\textbf{76.84\%}$ & $\textbf{94.55\%}$ \\
\hline 
& \multicolumn{3}{|c|}{15MCINC} & \multicolumn{3}{|c|}{30MCINC} \\
\hline
& ACC & SEN & SPE & ACC & SEN & SPE \\
\hline
MLDA &  $58.17\%$ & $53.03\%$ & $64.44\%$ & $62.34\%$ & $58.39\%$ & $66.06\%$ \\
\hline
SVM &  $61.00\%$ & $64.24\%$ & $57.04\%$ & $65.31\%$ & $54.19\%$ & $75.76\%$ \\
\hline
Lasso &  $53.83\%$ & $60.00\%$ & $46.30\%$ & $64.06\%$ & $59.68\%$ & $68.18\%$ \\
\hline
Elastic Net & $59.83\%$ & $64.85\%$ & $53.70\%$ & $65.78\%$ & $61.61\%$ & $69.70\%$ \\
\hline
Graphnet & $61.13\%$  & $64.85\%$ & $56.67\%$ & $66.09\%$  & $57.42\%$ & $74.24\%$ \\
\hline
TV + $\ell_{1}$ & $54.33\%$ & $60.71\%$  & $46.67\%$  & $61.56\%$ & $53.55\%$  & $69.09\%$ \\
\hline
$n^{2}$GFL & $60.50\%$ & $66.97\%$ & $52.59\%$ & $63.13\%$ & $61.61\%$ & $64.55\%$ \\
\hline
Ours($\beta_{\mathrm{pre}}$) &$\textbf{62.67\%}$  & $\textbf{68.18\%}$ & $\textbf{55.93\%}$ &$\textbf{67.03\%}$  & $\textbf{67.10\%}$ & $\textbf{66.97\%}$ \\
\hline
\end{tabular} 
\end{center}
\label{table:test}
\end{table}

 \subsection{Procedural Bias Captured by $\beta_{\mathrm{pre}}$}
\label{section-results}
\begin{figure}[!h]
\centering
\begin{minipage}[t]{0.9\linewidth}
   \includegraphics[width= \columnwidth]{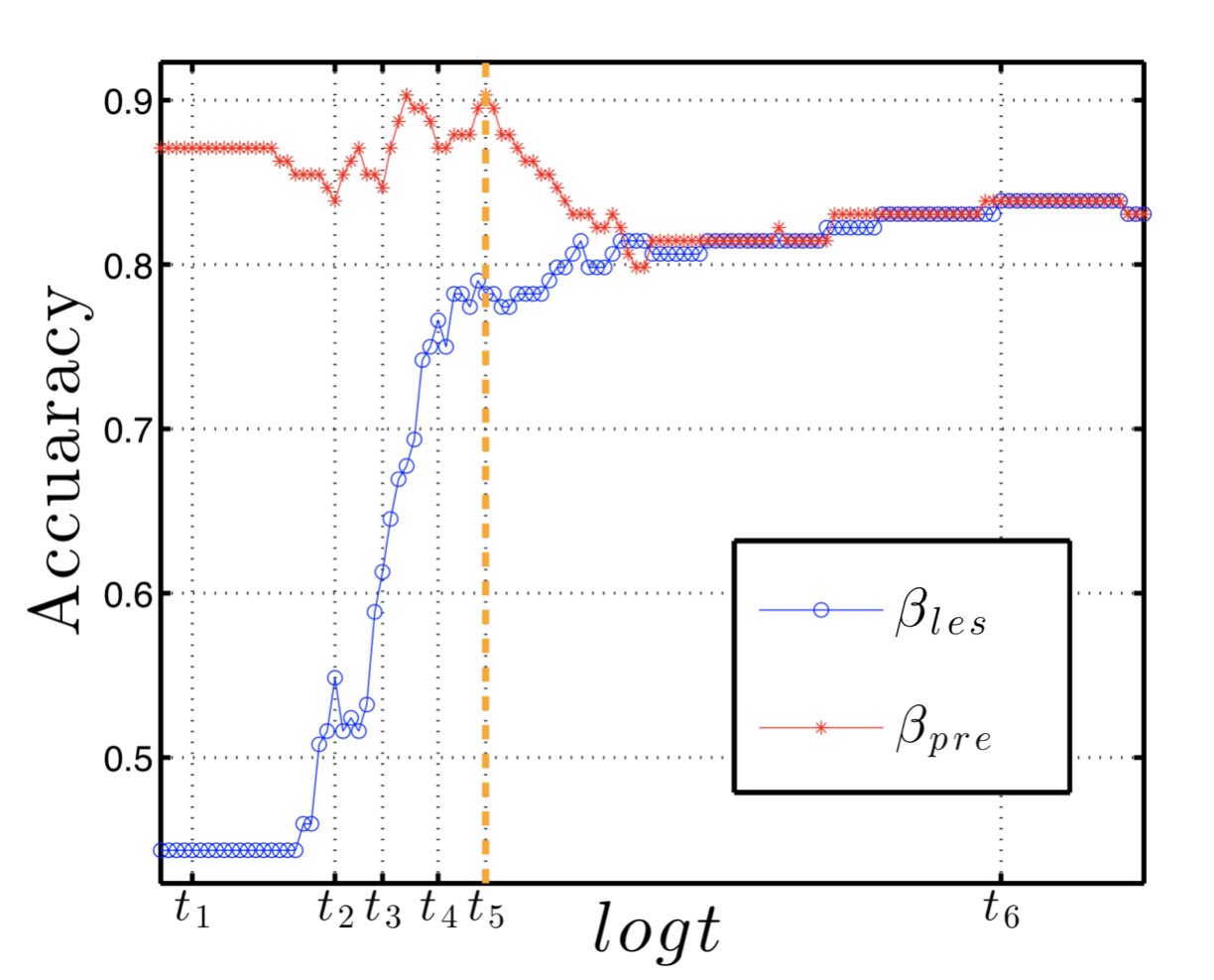}
\end{minipage}
\caption{Accuracy curve of $\beta_{\mathrm{pre}}$ and $\beta_{\mathrm{les}}$. Red curve represents $\beta_{\mathrm{pre}}$, blue curve represents $\beta_{\mathrm{les}}$.}
\label{figure:accuracy-curve}
\end{figure}

Recall that positive (negative) features represent degenerate (enlarged) voxels. By projecting $\beta_{\mathrm{pre}}$ onto the subspace of $\gamma$ (Eq.~\eqref{eq:prox}), $\beta_{\mathrm{les}}$ can be returned at each iteration. Note that with non-negative constraint, the $\beta_{\mathrm{les}}$ only select lesion features while $\beta_{\mathrm{pre}}$ has the capability of selecting other negative features to fit data better. To see such an effect of $\beta_{\mathrm{pre}}$ in comparison with $\beta_{\mathrm{les}}$, we plot accuracy curve on validation sets v.s. time $t$ due to the path property of our algorithm. The result of 30ADNC is used as an illustration. It can be seen in Fig.~\ref{figure:accuracy-curve} that (1) accuracy of $\beta_{\mathrm{pre}}$ is higher than $\beta_{\mathrm{les}}$ in the whole path; (2) before reaching its highest accuracy, i.e. $t_{6}$, $\beta_{\mathrm{pre}}$ shares the similar trend of accuracy with $\beta_{\mathrm{les}}$; after $t_{6}$, they gradually converge to each other when the GSplit LBISS starts to over-fit.

\begin{figure}[!h]
\centering
\begin{minipage}[t]{0.3\linewidth}
    \includegraphics[width= \columnwidth]{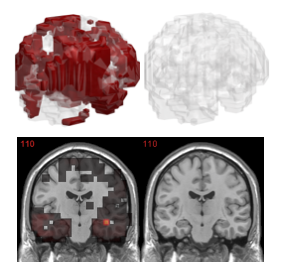}{\begin{center} $t = t_{1}$ \end{center}}
\end{minipage}
\begin{minipage}[t]{0.31\linewidth}
    \includegraphics[width= \columnwidth]{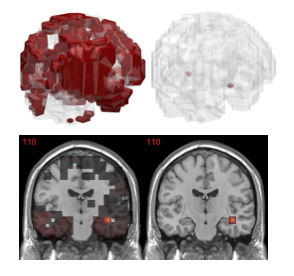}{\begin{center} $t = t_{2}$ \end{center}}
\end{minipage}
\begin{minipage}[t]{0.3\linewidth}
    \includegraphics[width= \columnwidth]{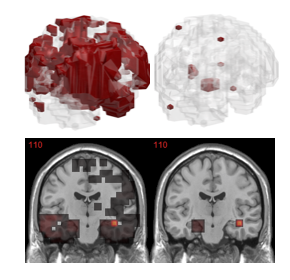}{\begin{center} $t = t_{3}$ \end{center}}
\end{minipage}
\begin{minipage}[t]{0.3\linewidth}
    \includegraphics[width= \columnwidth]{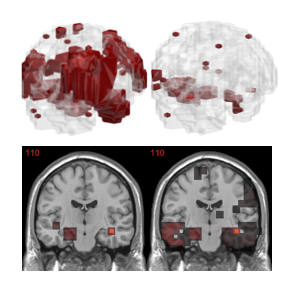}{\begin{center} $t = t_{4}$ \end{center}}
\end{minipage}
\begin{minipage}[t]{0.3\linewidth}
    \includegraphics[width= \columnwidth]{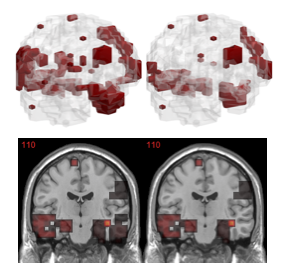}{\begin{center} $t = t_{5}$ \end{center}}
\end{minipage}
\begin{minipage}[t]{0.3\linewidth}
    \includegraphics[width= \columnwidth]{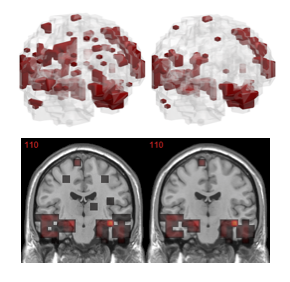}{\begin{center} $t = t_{6}$ \end{center}}
\end{minipage}
\caption{The 2-d 110 brain slices of $\beta_{\mathrm{les}}$ and $\beta_{\mathrm{pre}}$ at six time points above.}\label{figure:beta-pre-beta-les}
\end{figure}

To explain (2), we compare the positive (lesion) features selected by $\beta_{\mathrm{pre}}$ and $\beta_{\mathrm{les}}$ in six points during the regularization solution path. The 3-d images and 2-d slice images are also shown. The brighter of the color of the voxel, the larger of the value of it. We can see from Fig.~\ref{figure:beta-pre-beta-les} that $\beta_{\mathrm{les}}$ is more similar with $\beta_{\mathrm{pre}}$ as $t$ progresses. From the start (at $t_{1}$), almost all features are nonzeros and assigned with comparably large values for $\beta_{\mathrm{pre}}$ while no features selected for $\beta_{\mathrm{les}}$. As $\beta_{\mathrm{les}}$ continuously selects features located near 2-side hippocampus and thalamus on $t_{2}$, $t_{3}$, $t_{4}$, the corresponding features are brighter/larger than others in $\beta_{\mathrm{pre}}$ and the image looks more similar to $\beta_{\mathrm{les}}$. At $t_{5}$, $\beta_{\mathrm{pre}}$ captures features located in those atrophied areas from $\beta_{\mathrm{les}}$ and achieves the highest accuracy in the whole path. After $t_{5}$, the $\beta_{\mathrm{les}}$ starts to fit noise other than lesion features since with less regularization effect, leading to convergence of $\beta_{\mathrm{les}}$ to $\beta_{\mathrm{pre}}$. 

To explain (1), note that in addition to similar lesion features of $\beta_{\mathrm{pre}}$ and $\beta_{\mathrm{les}}$, 
$\beta_{\mathrm{pre}}$ can also capture additional procedural bias on other regions to achieve better fitness of data. Specifically, at $\beta_{\mathrm{pre}}$'s highest accuracy ($t_{5}$), the top 50 negative (enlargement) voxels of average $\beta_{\mathrm{pre}}$ are recorded, as shown in Fig.~\ref{figure:procedural-bias}. We can see that most selected voxels are located in GM voxels near lateral ventricle or subarachnoid space etc., which are different from locations of lesion ones.

\subsection{Gray Matter lesion feature analysis}
\label{section-lesion}
\begin{figure*}[t!]
\begin{center}

\begin{minipage}[t]{0.1370\linewidth}
    \includegraphics[width= \columnwidth]{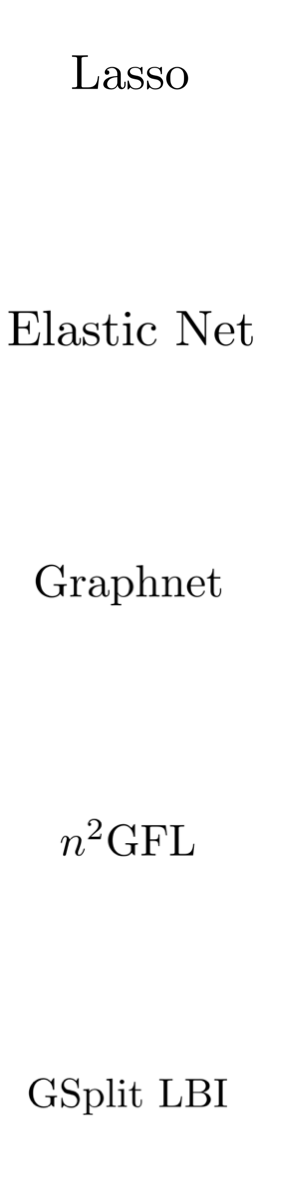}{\begin{center}  Model \end{center}}
 \end{minipage}  
\begin{minipage}[t]{0.1370\linewidth}
    \includegraphics[width= \columnwidth]{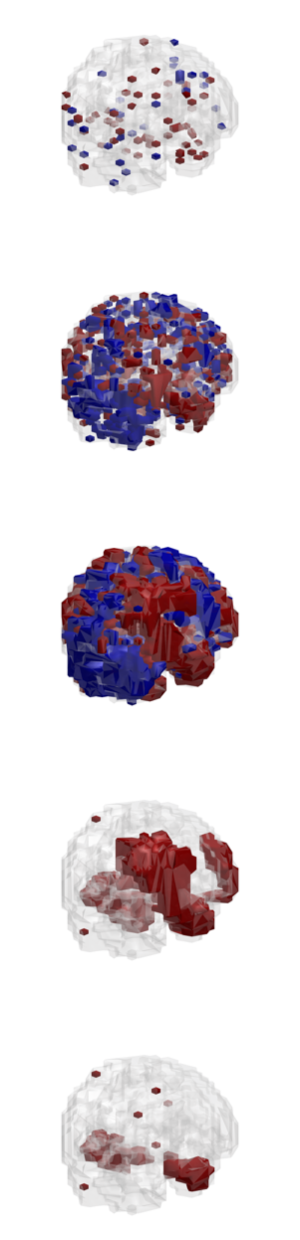}{\begin{center} (a) fold 1 \end{center}}
\end{minipage}
\begin{minipage}[t]{0.1370\linewidth}
    \includegraphics[width= \columnwidth]{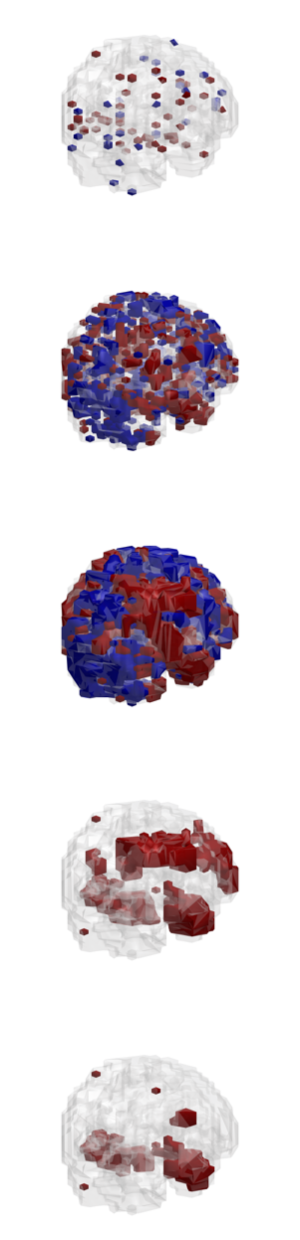}{\begin{center} (b) fold 3\end{center}}
    \end{minipage}
 \begin{minipage}[t]{0.1370\linewidth}
    \includegraphics[width= \columnwidth]{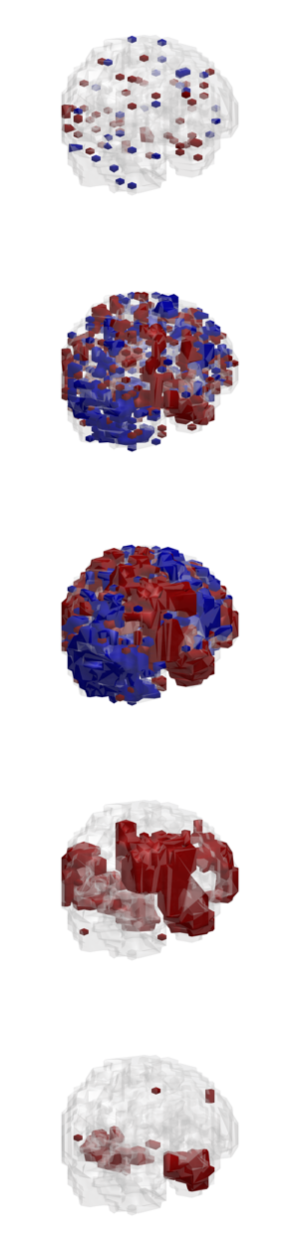}{\begin{center} (c) fold 5 \end{center}}
\end{minipage}
\begin{minipage}[t]{0.1370\linewidth}
    \includegraphics[width= \columnwidth]{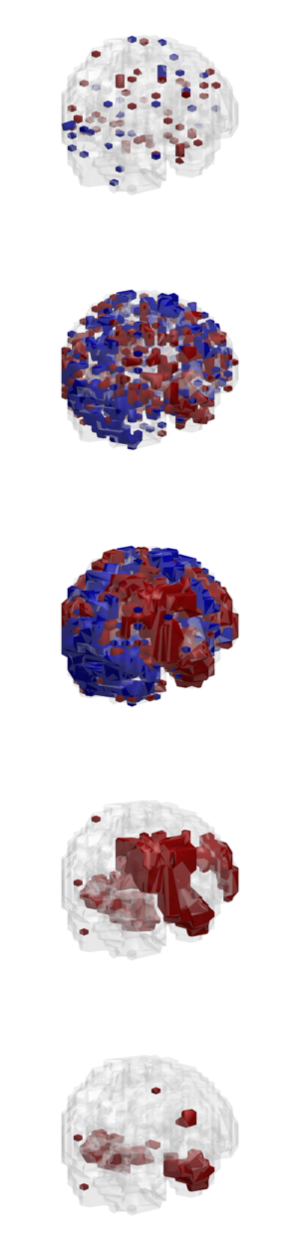}{\begin{center} (d) fold 7\end{center}}
    \end{minipage}
    \begin{minipage}[t]{0.1370\linewidth}
    \includegraphics[width= \columnwidth]{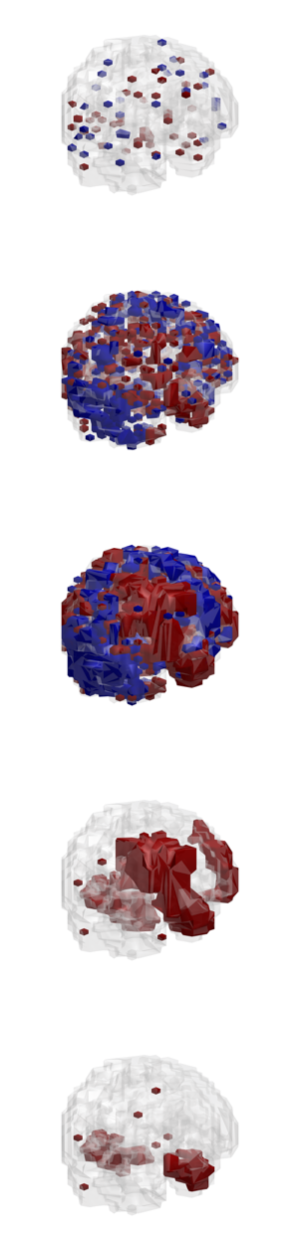}{\begin{center} (e) fold 9 \end{center}}
\end{minipage}
\begin{minipage}[t]{0.1370\linewidth}
    \includegraphics[width= \columnwidth]{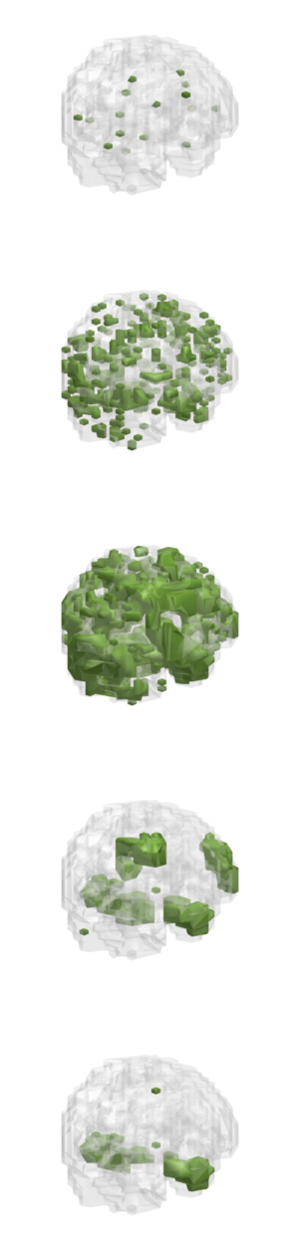}{\begin{center} (f) overlap\end{center}}
\end{minipage}
    \end{center}
    \caption{Selected lesion features across different folds by different models shown in 3-d brain images. The 1-5th row illustrated lasso, elastic net, graphnet, $n^2GFL$ and $\beta_{\mathrm{les}}$ of GSPlit LBI, respectively. (a)-(e): Results of fold 1,3,5,7,9. (f): The overlapped features in all 10 folds. }
\label{figure:lesion-3d}
    \end{figure*}

It's shown in Fig.~\ref{figure:illustrate} that by introducing variable splitting term, the tasks of prediction and selecting lesion features can be incorporated and solved by two different estimators. Hence, in addition to prediction power, the lesion feature analysis is another important component for evaluating our model. One of the main concern in such task is whether the selected lesion features are stable, which is important in terms of trustworthy to regard the lesion features as indicators in clinical diagnosis. To quantitatively evaluate such a stability, multi-set Dice Coefficient (mDC) is applied as a measurement and defined as:
\begin{equation*}
mDC := \frac{10 | \cap_{k=1}^{K} S(k) | }{\sum_{k=1}^{K} \thinspace | S(k) |} 
\label{mdc}
\end{equation*}
where $K$ is the number of folds in cross-validation and $S(k)$ denotes the support set of solution vector in k-th fold \cite{mdc,n2gfl}. We compute mDC for $\beta_{\mathrm{les}}^{k}$ at $t_{k}$ corresponding to the highest accuracy of $\beta_{\mathrm{les}}$ via 10-fold cross-validation on the whole dataset. The 30ADNC task is again applied as an example. As shown from Table~\ref{table:lesion}, when $\nu = 0.0004$ (corresponding to $\nu \to 0$ in Fig.~\ref{figure:illustrate}), the $\beta_{\mathrm{les}}$ of our model can obtain more stable lesion feature selection results than other models with comparable prediction power. Besides, the average number of selected features (line 3 in Table~\ref{table:lesion}) are also recorded. Note from Table~\ref{table:lesion} and Fig.~\ref{figure:lesion-3d} that although elastic net and graphnet (with $\ell_2$ regularization) are of slightly higher accuracy than $\beta_{\mathrm{les}}$, they select much more features than necessary. 

\begin{table*}[!h]
\caption{mDC comparisons between GSplit LBI and other models}
\begin{center}
\begin{tabular}{|c|c|c|c|c|c|c|}
\hline
 & Lasso & Elastic Net & Graphnet & $n^{2}$GFL & GSplit LBI\thinspace($\beta_{\mathrm{les}}$)\\
\hline
Accuracy & $87.50\%$ & $89.77\%$ & $90.34\%$ &  $87.50\%$ & $89.20\%$ \\
\hline
 mDC &0.1973 & 0.4524 & 0.6216 & 0.5362  & \textbf{0.7628} \\
\hline
$\sum_{k=1}^{10} \thinspace | S(k) |/10$ & 81.1 & 963.7 & 1,969.2 & 443.9 & 131.1  \\
\hline
\end{tabular} 
\end{center}
\label{table:lesion}
\end{table*}

\begin{figure*}[t!]
\begin{center}
\begin{minipage}[t]{0.15\linewidth}
    \includegraphics[width= \columnwidth]{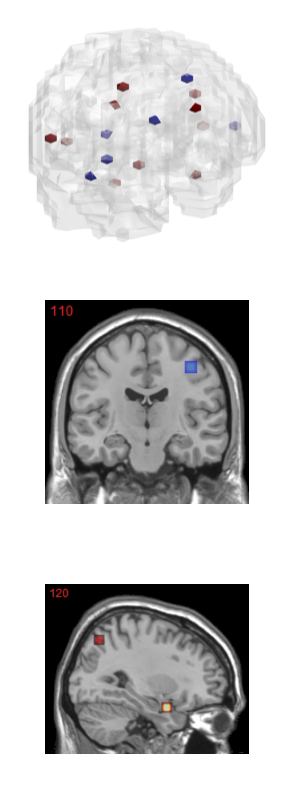}{\begin{center} (a) lasso \end{center}}
\end{minipage}
\begin{minipage}[t]{0.15\linewidth}
    \includegraphics[width= \columnwidth]{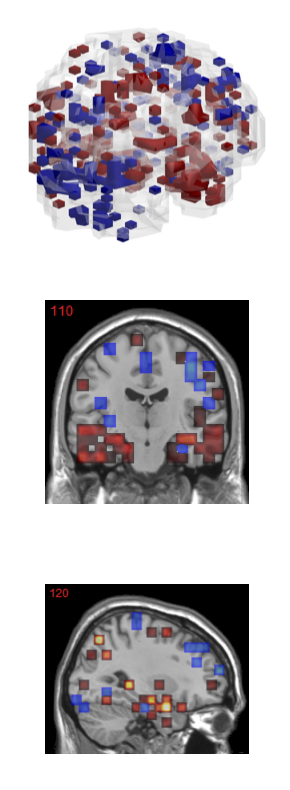}{\begin{center} (b) elastic\end{center}}
\end{minipage}
\begin{minipage}[t]{0.15\linewidth}
    \includegraphics[width= \columnwidth]{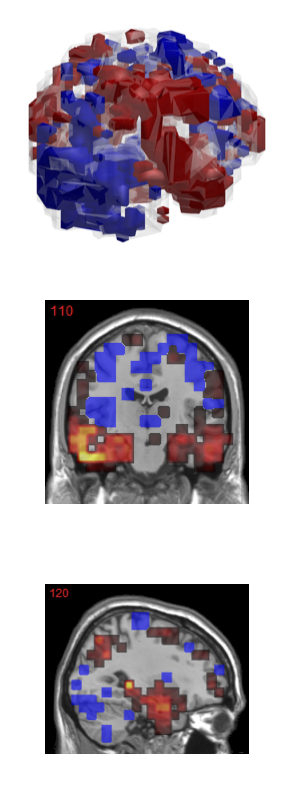}{\begin{center} (c) graphnet \end{center}}
\end{minipage}
\begin{minipage}[t]{0.15\linewidth}
    \includegraphics[width= \columnwidth]{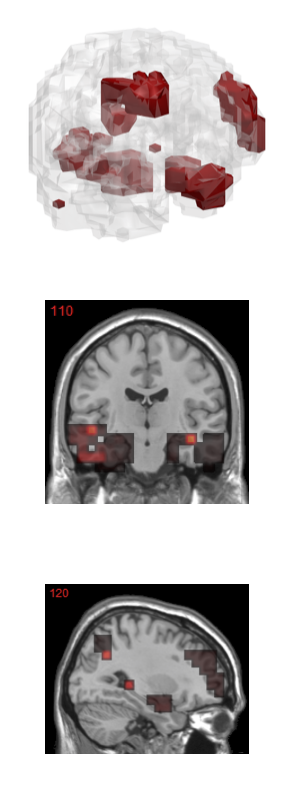}{\begin{center} (d) $n^2$GFL \end{center}}
\end{minipage}
\begin{minipage}[t]{0.15\linewidth}
    \includegraphics[width= \columnwidth]{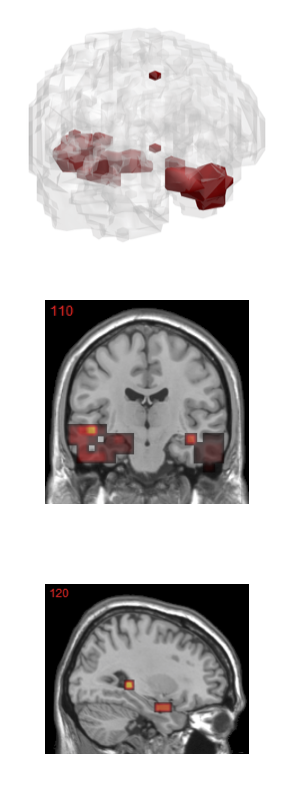}{\begin{center} (e) GSplit LBI ($\beta_{\mathrm{les}}$ \end{center}}
\end{minipage}
\end{center}
\caption{Overlapped features of different models shown in 2-d 110 brain slices. Red represents lesion features; blue represents procedural bias.}
\label{figure:lesion-2d}
\end{figure*}

In Fig.~\ref{figure:lesion-2d}, the overlapped features across all 10 folds of different models are illustrated (corresponding to their highest accuracy for cross-validation). We can see that, the features selected by lasso are too scattered to clustered into regions. Although the elastic net can select more correlated features, they do not form into cluster with no prior of 3D smooth. The features selected by graphnet are redundant (with an average of 1,969.2, among all 2,527 features), although they can form into regions. For $n^2$GFL and GSplit LBI, they consider non-negative, sparsity and 3D smooth sparsity requirements. As shown, the selected features are located in hippocampus, parahippocampal gyrus and medial temporal lobe etc., which are believed to be early damaged regions for AD patients. On the contrary, the other three models can select voxels that are located in other regions, which may be hard to explain the disease.

\subsection{Tuning parameters}

In our algorithm, parameter $\nu$ balances the prediction task and stability of feature selection, i.e. when $\nu \to 0$, the $\beta_{\mathrm{les}}$ can select features with better fitness of data and more stability; when $\nu$ gets larger, the $\beta_{\mathrm{pre}}$ will be assigned with more capability to select procedural bias. Therefore, the choice of $\nu$ is task-dependent. In this experiment, $\nu$ is set to 0.2 for prediction of disease and 0.0004 for lesion feature analysis.

The regularization parameter $t$ and parameter $\rho$ are determined via cross-validation. For $t$, we can see that on one hand $\beta_{\mathrm{pre}}$ will be constrained by $\beta_{\mathrm{les}}$ to continuously select more lesion features as $t$ progresses; on the other hand, there will be less capability of $\beta_{\mathrm{pre}}$ to capture procedural bias during the path since the regularization effect of $\beta_{\mathrm{les}}$ decreases at $t$ grows. At the optimal point, $\beta_{\mathrm{pre}}$ can appropriately "learn" lesion features from $\beta_{\mathrm{les}}$ and is not equal to $\beta_{\mathrm{les}}$ since it's also assigned with enough capability to utilize procedural bias to further improve prediction result compared to $\beta_{\mathrm{les}}$. When $\beta_{\mathrm{pre}}$ and $\beta_{\mathrm{les}}$ converge together, some redundant features can be selected by $\beta_{\mathrm{les}}$; besides, $\beta_{\mathrm{pre}}$ will lose the capability to capture procedural bias, in which case the accuracy will decrease.

Parameter $\rho$ can be treated as trade-off between geometric clustering and voxel sparsity, i.e. for larger value of $\rho$, the model tends to select more clustered features; for smaller value of $\rho$, features are tended to more scattered. Heuristically, the choice of $\rho$ depends on the quality of data, e.g. for data with lower resolution of MRI images, comparably larger $\rho$ is suggested since in this case more clustered features will be selected to suppress the noise. Take our experiments as an example, for tasks of 15ADNC and 15MCINC, the $\rho$ selected by cross-validation are 1.95 and 2.35, respectively; for tasks of 30ADNC and 30MCINC, the $\rho$ selected by cross-validation are 1.35 and 1.65, respectively. Such results agree with the common sense since normally 3.0T MRI scanner can produce images with higher resolution than 1.5T MRI scans.

\subsection{Coarse-to-fine experiments}

\begin{figure}[t!]
\begin{center}
    \includegraphics[width= 1.02\columnwidth]{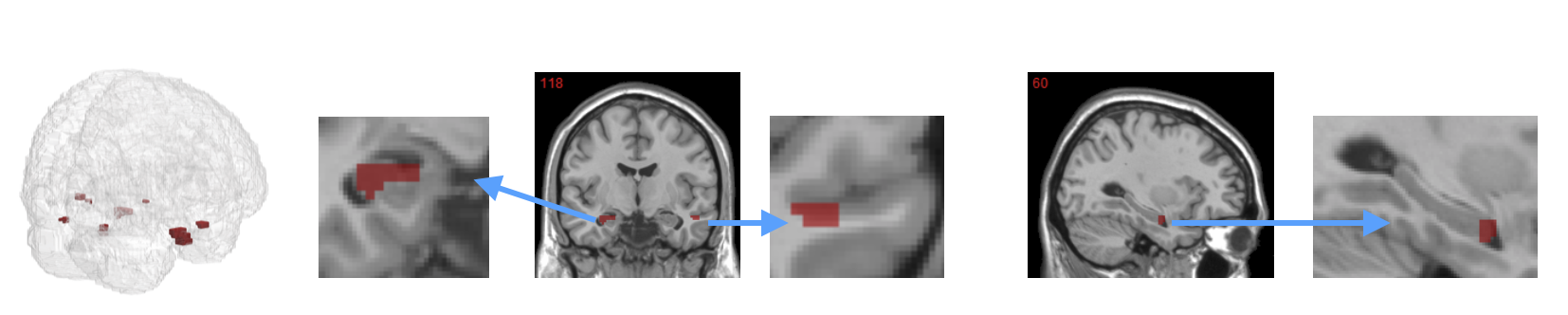}
\end{center}
\caption{The 2-d slice brain image of selected voxels with $2\times 2\times 2mm^{3}$ using coarse-to-fine approach.}
\label{figure:coarse-to-fine}
\end{figure}

To further investigate the locus of lesion features, an experiment with coarse-to-fine is conducted. Specifically, we project the overlapped voxels with $8 \times 8 \times 8mm^{3}$ size (shown in Fig.~\ref{figure:lesion-2d} (e)) onto MRI images with more finer scale voxels, i.e. $2 \times 2 \times 2 mm^{3}$ size. Totally 4,179 voxels are served as input features after projection. Again, the GSplit LBI is implemented using 10-fold cross validation. The accuracy on validation set of $\beta_{\mathrm{les}}$ is $90.34\%$ and on average 634.0 voxels are selected and they belong to parts of lesion regions, such as those located in Hippocampal Head, Medial Temporal Lobe, as shown in (d) of Fig.~\ref{figure:coarse-to-fine}.

\section{Conclusions}
\label{sec:discussion}

In this paper, a novel iterative variable splitting algorithm, based on differential inclusion of inverse scale space, is proposed to stably select lesion features and leverage procedural bias into prediction in neuroimage analysis. With a variable splitting term, the estimators for prediction and selecting lesion features can be separately pursued and mutually monitored under a gap control. The gap here is dominated by the procedural bias, some specific features crucial for prediction yet ignored in a priori disease knowledge. With experimental studies conducted on 15ADNC, 30ADNC, 15MCINC and 30MCINC tasks, we showed that: (1) the leverage of procedural bias can lead to significant improvements in both prediction and model interpretability, (2) the selected lesion features are with high stability and are located in regions that are believed to be early damaged. We believe that our methods can be extended to other applications. \newline

\newpage
\bibliographystyle{IEEEtran}
\bibliography{IEEEabrv,tmi_gsplitlbi}

%end{thebibliography}


\begin{comment}
\begin{IEEEbiography}{Michael Shell}
Biography text here.
\end{IEEEbiography}

\begin{IEEEbiographynophoto}{John Doe}
Biography text here.
\end{IEEEbiographynophoto}

%\newpage

\begin{IEEEbiographynophoto}{Jane Doe}
Biography text here.
\end{IEEEbiographynophoto}
\end{comment}

%\vfill


\end{document}